\documentclass[journal]{IEEEtran}

\usepackage{epsfig} %
\usepackage[cmex10]{amsmath} %
\usepackage{amssymb}  %
\usepackage{amsthm}
\usepackage{graphicx}
\usepackage{array}
\usepackage{cite}
\usepackage{subcaption}
\usepackage{algorithm}
\usepackage{algpseudocode}
\usepackage{xcolor} %
\usepackage{siunitx} %
\usepackage{booktabs} %
\usepackage{multirow} %
\usepackage{url}

\DeclareCaptionLabelSeparator{ieeefigcapspacing}{.\nobreakspace\nobreakspace}
\DeclareCaptionLabelSeparator{ieeetblcapspacing}{\\}
\DeclareMathOperator*{\argmin}{\arg\min}
\providecommand{\norm}[1]{\lVert#1\rVert}
\providecommand{\abs}[1]{\lvert#1\rvert}
\providecommand{\reals}[1]{\mathbb{R}^{#1}}
\providecommand{\mat}{}
\renewcommand{\vec}{}
\providecommand{\nats}[1]{\mathbb{N}^{#1}}

\newtheorem{theorem}{Theorem}
\newtheorem{corollary}{Corollary}
\newtheorem{proposition}{Proposition}
\newtheorem{assumption}{Assumption}
\newtheorem{definition}{Definition}

\newfloat{algcap}{h}{lop}
\floatname{algcap}{Algorithm}

\usepackage{soul}
\renewcommand{\hl}[1]{{#1}}
\newcommand{\edit}[1]{{#1}}
\newenvironment{newsection}{}{\ignorespacesafterend}
\newcommand{\new}[1]{{#1}}
\newenvironment{newersection}{}{\ignorespacesafterend}
\newcommand{\newer}[1]{{#1}}

\begin{document}

\title{Sequential Action Control: Closed-Form Optimal Control for Nonlinear and Nonsmooth Systems}

\author{Alex~Ansari and~Todd~Murphey%
\thanks{Manuscript received October 27, 2015; revised June 4, 2016; accepted July 9, 2015.}%
\thanks{A. Ansari is with the Robotics Institute, Carnegie Mellon University, Pittsburgh, PA 15213 USA (e-mail: {\tt\footnotesize aansari1@andrew.cmu.edu}).}%
\thanks{T. Murphey is with the Department of Mechanical Engineering, Northwestern University, Evanston, IL 60208 (e-mail: {\tt\footnotesize t-murphey@northwestern.edu}).}}


%
%
%
%

\maketitle

\begin{abstract}

This paper presents a new model-based algorithm that computes predictive optimal controls on-line and in closed loop for traditionally challenging nonlinear systems.  Examples demonstrate the same algorithm controlling hybrid impulsive, underactuated, and constrained systems using only high-level models and trajectory goals.  
Rather than iteratively optimize finite horizon control sequences to minimize an objective, this paper derives a closed-form expression for individual control actions, i.e., control values that can be applied for short duration, that optimally improve a tracking objective over a long time horizon.
Under mild assumptions, actions become linear feedback laws near equilibria that permit stability analysis and performance-based parameter selection.  
Globally, optimal actions are guaranteed existence and uniqueness. 
By sequencing these actions on-line, in receding horizon fashion, the proposed controller provides a min-max constrained response to state that avoids the overhead typically required to impose control constraints. 
Benchmark examples show the approach can avoid local minima and outperform nonlinear optimal controllers and recent, case-specific methods in terms of tracking performance, and at speeds orders of magnitude faster than traditionally achievable.

\end{abstract}

\begin{IEEEkeywords}
real-time optimal control; nonlinear control systems; hybrid systems; impacting systems; closed loop systems.
\end{IEEEkeywords}

\section{INTRODUCTION}
\label{intro}

\begin{figure}[t!]
  \vspace{-.15in}
  \centering
\includegraphics[width=3.2in]{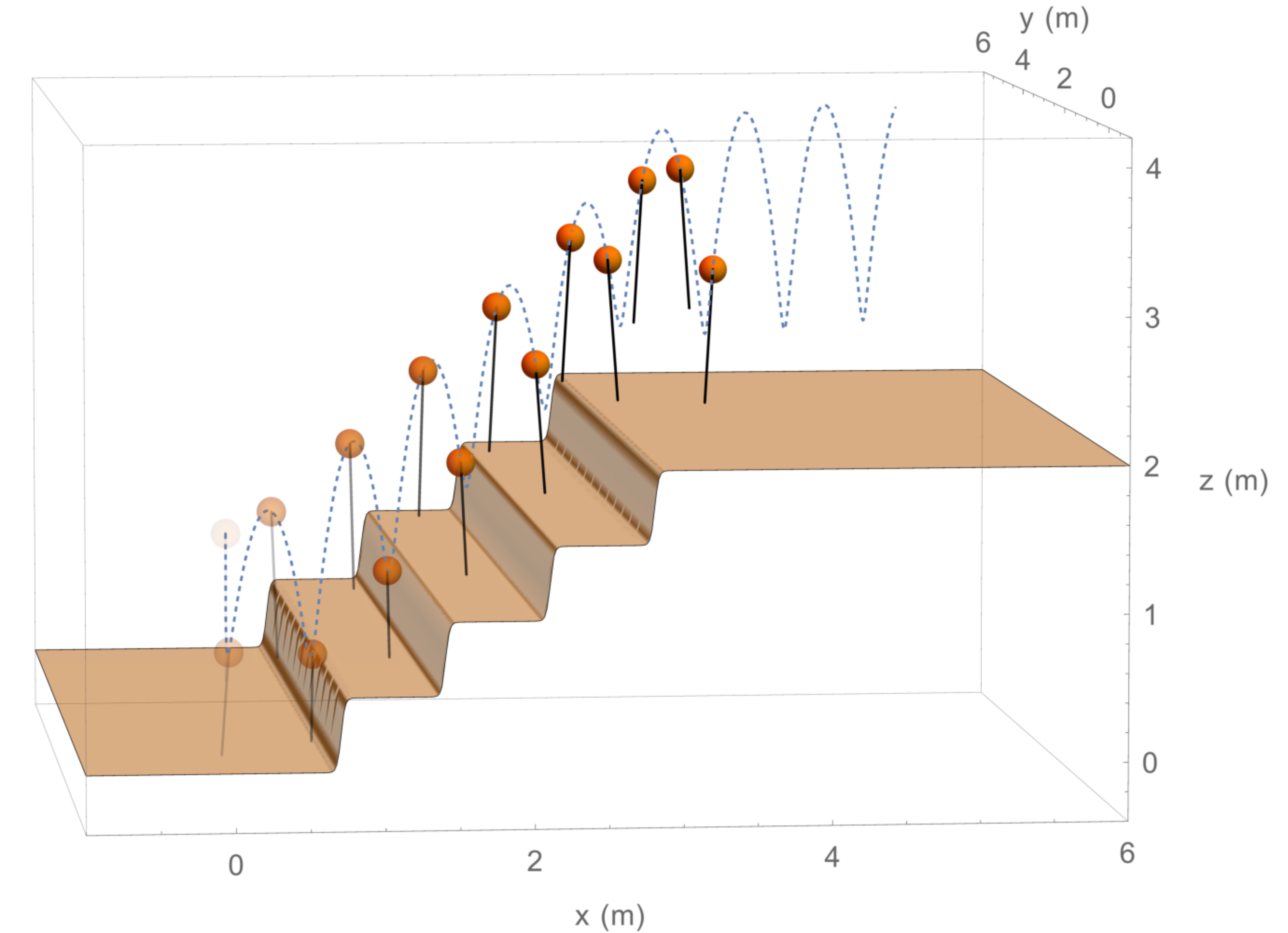}
  \caption{\newer{Time lapse (0.5s) of a spring-loaded inverted pendulum (SLIP) reactively hopping up stairs using SAC.}}
  \label{fig:intro_slip_hopping}
\end{figure}

\IEEEPARstart{M}{odel-based} control techniques like dynamic programming or trajectory optimization can generate highly efficient motions that leverage, rather than fight, the dynamics of robotic mechanisms.
Yet, these tools are often difficult to apply since even basic robot locomotion and manipulation tasks often yield complex nonlinear, hybrid, underactuated, and high-dimensional constrained control problems. For instance, consider the spring-loaded inverted pendulum (SLIP) model in Fig.~\ref{fig:intro_slip_hopping}.  With only a point-mass body and spring leg, the model provides one of the most idealized locomotion templates that remains popular in robotics for capturing the center of mass (COM) running dynamics of a wide range of species \cite{SpringMassWalkRun2006,NaturePendWalkRunModel2005}.  However, even this simple model exhibits nonlinear, underactuated, and hybrid phenomena that affect optimizations in most model-based methods.

For example, a trajectory optimization approach could derive the dashed solution in Fig.~\ref{fig:intro_slip_hopping}, but would usually require a good initial guess (and special care to account for discontinuities). Initialization is important because the nonlinear dynamics imply the constrained objective is non-convex and subject to potentially poor local minima.  An additional regulating controller, capable of tracking the trajectory through impacts, would also be necessary to track the resulting solution.  Still, the approach would not be able to adapt the trajectory to accommodate a dynamically changing environment.

As an alternative, a (nonlinear) receding horizon control approach would compute an optimal trajectory, follow it for a single time step, and then iterate to construct a closed-loop response. However, each receding horizon problem would still be non-convex, requiring computationally expensive iterative optimization \cite{NMPCbook2000}.  In addition to local minima issues, the approach would be limited to lower bandwidth scenarios with slowly-varying environmental/dynamic conditions \cite{NMPC2004,NMPCbook2000}.

\IEEEpubidadjcol

\begin{figure}[t!]
\vspace{-.1in}
  \centering
  \includegraphics[width=2.6in]{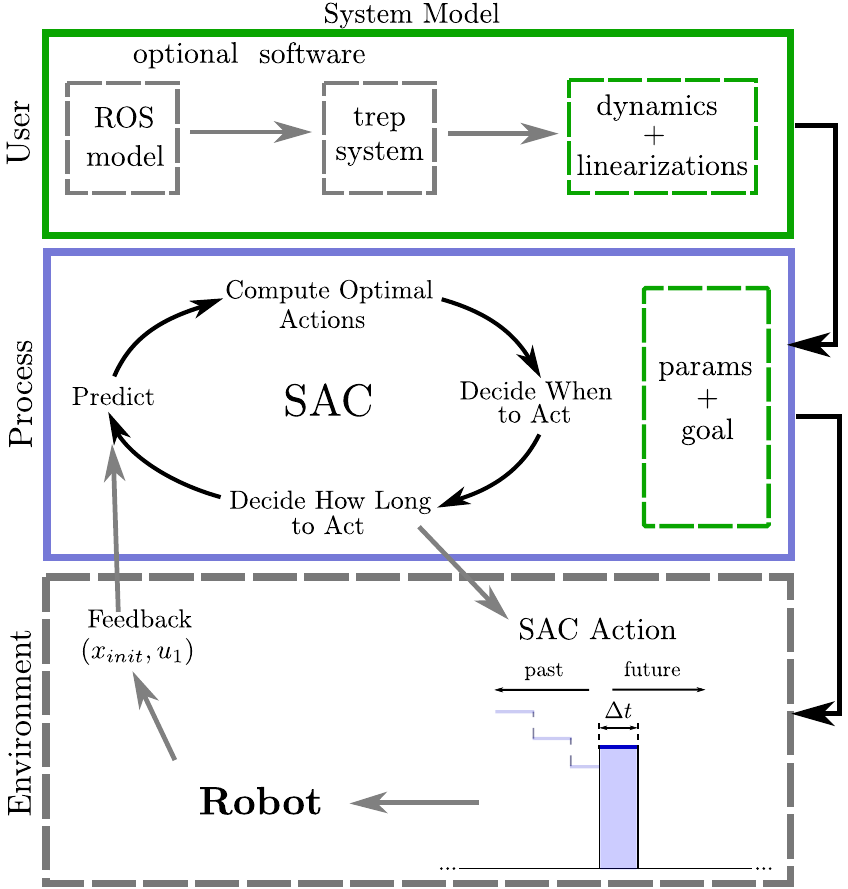}
  \caption{\new{An overview of the SAC control process including possible open-source interfaces, e.g., ROS \cite{ros} and trep \cite{trep}.%
}}
  \label{fig:intro_sac_process}
\end{figure}

To address these problems, 
this paper proposes a new model-based algorithm, which we refer to as \emph{Sequential Action Control} (SAC), that makes strategic trade-offs for computational gain and improved generality.
That is, rather than solving for full control curves that minimize a non-convex objective over each receding horizon, SAC finds a single optimal control \emph{value} and \emph{time} to act that maximally \emph{improves} performance.\footnote{\newer{SAC also uses a line search \cite{NocedalOptimization2006} to specify a short duration (usually a single discrete time-step) to apply each control and improve performance.}}  For instance, in the SLIP example SAC waits until the flight phase to tilt the leg backwards, predicting that the action will drive the robot farther up the steps (after the leg re-contacts) for some specified horizon. As in receding horizon control, SAC incorporates feedback and repeats these calculations at each time step as the horizon recedes.  The resulting process computes a real-time, closed-loop response that reacts to terrain and continually drives the robot up the steps. Figure~\ref{fig:intro_sac_process} provides an overview of the SAC process.

There are several advantages to the trade-offs SAC makes, i.e., computing individual control actions at each time step that improve performance rather than curves that directly optimize a performance objective.
These advantages include: 1) SAC controls can be rapidly computed on-line from a closed-form expression with guaranteed optimality, existence, and uniqueness.\footnote{There are algorithms other than SAC that yield controls in closed form.  However, we are unaware of any methods (particularly model/optimization-based methods) that provide comparable constrained closed-form controls on-line for examples such as those in Section~\ref{examples} and accommodate nonlinear hybrid/impulsive robots.}
 2) SAC controls can be directly saturated to obey min-max constraints without any computational overhead or specialized solvers. 3) SAC's control synthesis process is unaffected by discontinuities in dynamics and so applies to challenging hybrid and impulsive robots.
4) In spite of sacrificing the multi-step planning process of trajectory optimization, benchmark examples demonstrate a final, unintuitive finding -- SAC can avoid poor local minima that trap nonlinear optimal control.  
To illustrate this last point, Section~\ref{examples} includes a number of robotics-related control examples that show SAC outperforms case-specific methods and popular optimal control algorithms (sequential quadratic programming \cite{NocedalOptimization2006} and iLQG \cite{iLQGcontConst2014}).  Compared to these alternatives, SAC computes high-bandwidth ($1$ KHz) closed-loop trajectories with equivalent or better final cost in less time (milliseconds/seconds vs. hours).

To sum up, SAC provides a model-based control response to state that is easily implemented and efficiently computed for for most robotic systems, including those that are saturated, underactuated, nonlinear, hybrid/impulsive, and high dimensional.  This paper introduces SAC in two parts. Part I focuses on robots with differentiable nonlinear dynamics, and Part II considers hybrid impulsive robots.  Benchmark examples and relevant background material are introduced in the context of each.
Table~\ref{table:notation} includes notation used throughout this paper.

\begin{figure}[t]
\centering
\includegraphics[width=2.65in]{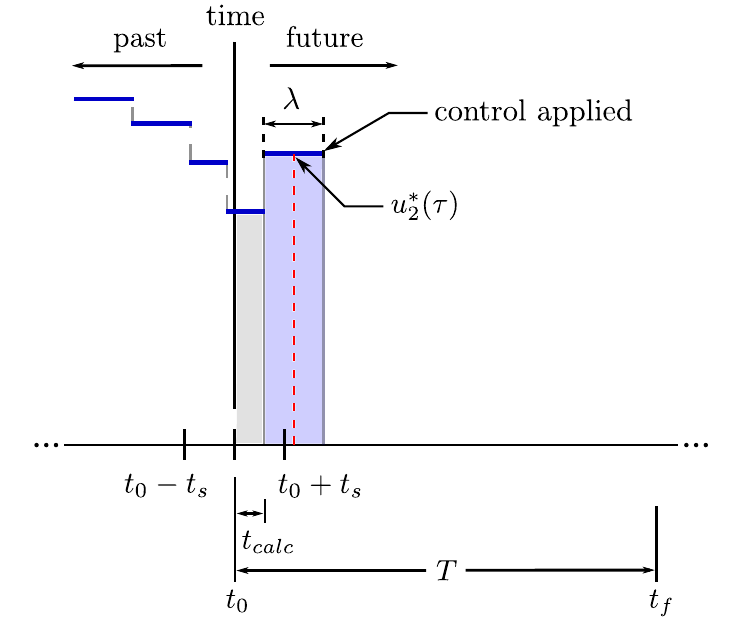}
\caption{\new{Following the cyclic process in Fig.~\ref{fig:intro_sac_process}, SAC computes a \emph{schedule}, $\vec u_2^{\,*} : (t_0,t_f) \mapsto \reals{m}$, providing the value of \newer{optimal actions that} maximally improve a tracking objective over the current (receding) horizon.
\newer{Next, SAC selects an application time, $\tau \in (t_0+t_{calc},t_f)$, and $\vec u_2^{\,*}(\tau)$ becomes the value of the next SAC action (blue shaded bar).}
A line search sets the duration, $\lambda$.  Previously computed actions are applied while current calculations complete, $t \in [t_0, t_0+t_{calc}]$.  \newer{After incorporating feedback, SAC repeats the process at the next sample time, $t = t_0 + t_s$.}}}
\label{fig:tscales}
\end{figure}

\section*{\new{Part I: SAC for Differentiable Systems}}
\section{Control Synthesis}
\label{control}

\begin{table}[h]
\captionsetup{font=footnotesize, labelsep=ieeetblcapspacing}
\renewcommand{\arraystretch}{1.4}
\centering
\caption{\scshape{Notation}}
\label{table:notation}
\begin{tabular}{ll}
\hline
symbol & description\\
\hline
$D_x f(\cdot)$ & partial derivative $\frac{\partial f (\cdot)}{\partial x}$\rule{0pt}{1.2em}\\
$\norm{\cdot}_{\mat M}$ & norm where $\mat M$ provides the metric\\
                      & (e.g., $\norm{\vec x(t)}_{\mat Q}^2 = \vec x(t)^T \mat Q \, \vec x(t)$\,)\\
$\mat{R}^{-T}$ & equivalent to $(\mat{R}^T)^{-1}$\\
$\mat{R} > \mat{0}$ & indicates $\mat R$ is positive definite ($\geq$ for semi-definite)\rule[-2.5mm]{0pt}{0pt}\\
\hline
\end{tabular}
\end{table}

The subsequent sections detail SAC control synthesis, following the cyclic process in Fig.~\ref{fig:intro_sac_process}.  
We describe how each cycle of the SAC process computes an optimal \emph{action} -- \emph{defined by the triplet consisting of a control's value, $\vec u \in \reals{m}$, a short application duration, $\lambda \in \reals{+}$, and application time, $\tau \in \,\newer{\reals{+}}$ (see the blue shaded bar in Fig.~\ref{fig:tscales})} -- that is sent to a robot.

\subsection{Prediction}
\label{sec:prediction}

At fixed sampling times every $t_s$ seconds, SAC measures the current (initial) state, $x_{init}$, and begins control synthesis by predicting the nominal motion of a robotic system over a receding horizon.  
Prediction starts at the current (initial) time, $t_0$, and extends to final time, $t_f$, with a horizon length, $T = t_f - t_0$.
So, for example, SAC produces the $10\,$s SLIP trajectory in Fig.~\ref{fig:intro_slip_hopping} by cycling through a synthesis process every $t_s = 0.01\,$s ($100$ Hz), with each prediction phase lasting $T=0.6\,$s. 
In each cycle, SAC computes an action that improves the $0.6\,$s predicted trajectory.  Repeating the process, SAC generates a piecewise continuous response.

In Part I, the dynamics,
\begin{equation}
\label{xdot}
\dot{\vec x}(t) = \vec f(t,\vec x(t),\vec u(t)) \text{\,,}
\end{equation}
are nonlinear in state $\vec x : \reals{} \mapsto \reals{n}$. Though these methods apply more broadly, we derive controls for the case where \eqref{xdot} is linear with respect to the control, $\vec u : \reals{} \mapsto \reals{m}$, satisfying control-affine form,
\begin{equation}
\label{f}
\vec f(t,\vec x(t),\vec u(t)) = \vec g(t,\vec x(t)) + \mat h(t,\vec x(t)) \, \vec u(t) \text{\,.}
\end{equation}
The time dependence in \eqref{xdot} and \eqref{f} will be dropped for brevity.

The prediction phase simulates motion resulting from some choice of nominal control, $u = u_1$.  Thus, the nominal predicted motion corresponds to,
\[
\new{\vec f_1 \triangleq \vec f(\vec x(t),\vec u_1(t)) \text{.}}
\]
Although the nominal control may be chosen arbitrarily, all examples here use a null nominal control, $u_1 = 0$.  
Hence, in the SLIP example, SAC seeks actions that improve performance relative to doing nothing, i.e., letting the SLIP fall. 

With $l_1: \reals{n} \mapsto \reals{}$ and $m : \reals{n} \mapsto \reals{}$, 
the cost functional, 
\begin{equation}
\label{J}
J_1 { }={ } \int_{t_0}^{t_f} \new{l_1(\vec x(t))} \,dt + m(\vec x(t_f))\text{\,,}
\end{equation}
measures trajectory performance to gauge the improvement provided by SAC actions.\footnote{Though not required, \eqref{J} should be non-negative if it is to provide a performance measure in the formal sense.}  
The following assumptions further clarify the \new{systems and cost functionals addressed.}

\begin{assumption}
\label{assump:dynam}
The elements of the dynamics, $f(t,x(t),u(t))$, are real, bounded, $\mathcal{C}^1$ in $\vec{x}(t)$, and $\mathcal{C}^0$ in $t$ and $\vec{u}(t)$.
\end{assumption}

\begin{assumption}
\label{assump:cost}
The terminal cost, $m(\vec x(t_f))$, is real and differentiable with respect to $\vec{x}(t_f)$.  Incremental cost \new{$l_1(\vec x(t))$} is real, Lebesgue integrable, and $\mathcal{C}^1$ \new{in $\vec{x(t)}$.}
\end{assumption}

\noindent The prediction phase concludes with simulation of \eqref{xdot} and \eqref{J}.

\subsection{Computing Optimal Actions}
\label{sec:actions}

Since SAC has not yet decided \emph{when} to act, it derives a schedule (curve), $\vec u_2^{\,*} : (t_0,t_f) \mapsto \reals{m}$, providing the value of the optimal action at every moment along the predicted motion.  
For instance, in the SLIP example SAC may determine that 1 N-m of torque at the current time will tilt the leg backwards sufficiently for the SLIP to bounce forward and improve \eqref{J}.  The same strategy may be optimal at a later time, e.g., just before leg touchdown, but require 10 N-m to accelerate the leg into position before impact.
In this scenario, $\vec u_2^{\,*}$ would provide these optimal torque values at each time and SAC would choose one ``best'' action to take.
At every sample time, SAC would update $\vec u_2^{\,*}$ and choose another action.

Modeling a SAC action as a short perturbation in the predicted trajectory's nominal control, this section derives the optimal action to apply at a given time by finding the perturbation that optimizes trajectory improvement.
Given the application time, $\tau \in (t_0,t_f)$, a (short) duration, $\lambda$, and the optimal action value, $\vec u_2^{\,*}(\tau)$, the perturbed control signal is piecewise continuous based on Def.~\ref{def:piecewise} and Assump.~\ref{assump:control}.
\begin{definition}
\label{def:piecewise}
\new{\edit{Piecewise continuous functions will be referred to as $\widetilde{\mathcal{C}}^0$.  These functions will be defined according to one of their one-sided limits at discontinuities.}}
\end{definition}
\begin{assumption}
\label{assump:control}
SAC control signals, $\vec{u}$, are real, bounded, and $\widetilde{\mathcal{C}}^0$ such that 
\begin{displaymath}
   \vec{u}(t) = \left\{
     \begin{array}{lr}
       \vec u_1(t) & : t \notin [\tau-\frac{\lambda}{2}, \tau+\frac{\lambda}{2}]\\
       \vec{u}_2^{\,*}(\tau) & : t \in [\tau-\frac{\lambda}{2}, \tau+\frac{\lambda}{2}] 
     \end{array}
   \right.\text{\,,}
\end{displaymath}
with nominal control, $\vec u_1$, that is $\mathcal{C}^0$ in $t$.\footnote{The dynamics and nominal control can be $\widetilde{\mathcal{C}}^0$ in $t$ if application times, $\tau$, exclude points of discontinuity in $\vec u_1(t)$.}
\end{assumption}
\noindent Hence, over each receding horizon, SAC assumes the system evolves according to nominal dynamics, $f_1$, except for a brief duration, where it switches to the alternate mode,
\[
\vec f_2 \triangleq  f(\vec x(t),\vec{u}_2^{\,*}(\tau))\text{.}
\]
SAC seeks the vector $\vec{u}_2^{\,*}(\tau)$ that optimally improves cost \eqref{J}.

In Part II, we derive a local model of the change in cost resulting from the perturbed SAC control signal and solve for actions that optimize improvement.  
In the present case of differentiable dynamics \eqref{xdot} however, the local model of the change in cost corresponds to an existing term from mode scheduling literature.
That is, we can re-interpret the problem of finding the change in cost \eqref{J} due to short application of $\vec{u}_2^{\,*}(\tau)$, as one of finding the change in cost due to inserting a new dynamic mode, $f_2$, into the nominal trajectory for a short duration around $t=\tau \in (t_0,t_f)$.
In this case, the \emph{mode insertion gradient} \cite{MagnusModeInsertion,MagnustInsertionGradientTransitionTime},
\begin{equation}
\label{dJdlambda}
\frac{dJ_1}{d \lambda^+}(\tau,u_2^{\,*}(\tau)) = \vec \rho(\tau)^T\bigg[\vec f(x(\tau),u_2^{\,*}(\tau)) -\vec f(x(\tau),u_1(\tau))\bigg]\text{\,,}
\end{equation}
provides a first-order model of the change in cost \eqref{J} relative to the duration of mode $f_2$.  The model is local to the neighborhood where the duration of the switch to/from $f_2$ approaches zero, $\lambda \rightarrow 0^+$.
Note that \eqref{dJdlambda} assumes the state in $\vec f_1$ and $\vec f_2$ is defined from the nominal control and $\vec \rho : \reals{} \mapsto \reals{n}$ is the adjoint variable calculated from the nominal trajectory,\footnote{As opposed to traditional fixed-horizon optimal control methods \cite{liberzonOptControl2012,PontryaginOptProcesses1962}, this adjoint is easily computed because it does not depend on the closed-loop, optimal state $\vec{x}^{*}(t, \vec{u}_2^{\,*}(\tau))$.}
\begin{equation}
\label{rhodot}
\dot{\vec \rho} = -\nabla l_1(\new{\vec x}) - D_{\vec x} \vec f(\vec x, \vec u_1)^{T} \vec \rho\text{\,,}
\end{equation}
with $\vec \rho(t_f) = \nabla \, m(\vec x(t_f))$.

\newer{
The mode insertion gradient is typically used in mode scheduling \cite{TimSuffDescentModeSched2016, MagnustInsertionGradientTransitionTime, MagnusInsertionGradient, MagnusInsertionGradientDeriv} to determine the optimal time to insert control modes assuming the modes are known a priori.  In this section, we use the mode insertion gradient to solve for new optimal modes (optimal actions) at each instant.\footnote{\newer{Also, Section~\ref{hybrid_sac} shows a local hybrid cost model yields a generalized version of \eqref{dJdlambda} for hybrid impulsive dynamical systems with resets and objectives that depend on the control.  A discussion is in Appendix~\ref{sec:mode_insert_grad}}.}
}

One way to interpret the mode insertion gradient is as a sensitivity.  
That is, 
the mode insertion gradient \eqref{dJdlambda} indicates the sensitivity
of cost \eqref{J} to an action's application duration at any potential application time, $\tau \in (t_0,t_f)$.  To achieve a desired degree of cost improvement with each action, SAC uses a control objective to select optimal actions that drive the cost sensitivity \eqref{dJdlambda} toward a desired negative value, $\alpha_d \in \reals{-}$.
At any potential application time $\tau \in (t_0,t_f)$, the action value, $u_2(\tau)$, that minimizes,
\newer{\begin{equation}
\label{l2}
l_2(\tau\newer{,u_2(\tau)}) \triangleq \frac{1}{2} [\frac{dJ_1}{d \lambda^+}(\tau,u_2(\tau)) - \alpha_d]^2 + \frac{1}{2} \norm{\vec{u}_2(\tau)}_{\mat R}^2 \text{\,,}
\end{equation}}\noindent
minimizes control authority in achieving the desired sensitivity.  %
\newer{The} matrix $\mat{R} = \mat{R}^T > \mat{0}$ provides a metric on control effort. Because the space of positive semi-definite / definite cones is convex \cite{BoydConvexOpt}, \eqref{l2} is convex with respect to action values, $\vec u_2(\tau)$.

\newer{With Assumps.~\ref{assump:dynam}-\ref{assump:control}, the mode insertion gradient exists, is bounded, and \eqref{l2} can be minimized with respect to $\vec{u}_2(\tau) \; \forall \tau \in (t_0,t_f)$.  
The following theorem, which stems from early work in \cite{CORacc2015}, finds this minimum to compute the schedule of optimal action values.}

\begin{theorem}
\label{uthrm}
Define $\mat \Lambda \triangleq \mat h(\vec x)^T \, \vec \rho \, \vec \rho^{\,T} \, \mat h(\vec x)$.  The schedule \newer{providing the value of the optimal action,
\begin{equation}
\label{u2sched}
\vec{u}_2^{\,*}(t) \triangleq \argmin_{\vec{u}_2(t)} \;l_2(t,u_2(t)) \;\;\;\;\;\; \forall t \in (t_0,t_f) \text{\,,}
\end{equation}}
to which cost \eqref{J} is optimally sensitive at any time is
\setlength{\arraycolsep}{0.0em}
\begin{eqnarray}
\label{U2Opt}
\vec u_2^{\,*} { }=&{ }& \; (\mat \Lambda + \mat{R}^T)^{-1} \, [\mat \Lambda \, \vec u_1 + \mat h(\vec x)^T \vec \rho \, \alpha_d] \text{\,.}%
\end{eqnarray}
\setlength{\arraycolsep}{5pt}
\end{theorem}

\begin{IEEEproof}
Evaluated at any time $t \in (t_0,t_f)$, \newer{$\vec u_2^{\,*}$ provides the value of the optimal action} that minimizes \eqref{l2} at that time.  The schedule therefore also minimizes the (infinite) sum of costs \eqref{l2} associated with the \newer{optimal action values} at every time $\forall t \in (t_0,t_f)$.  Hence, \eqref{u2sched} can be obtained by minimizing
\newer{\begin{equation}
\label{J2}
J_2 { }={ } \int_{t_0}^{t_f} l_2(t,\vec u_2(t)) \,dt\text{\,.}
\end{equation}}\noindent
\indent Because the sum of convex functions is convex, and $\vec x$ in \eqref{dJdlambda} depends only on $\vec{u}_1$, minimizing \eqref{J2} with respect to $\vec u_2(t) \, \forall t \in (t_0, t_f)$ is convex and unconstrained.  It is necessary and sufficient for (global) optimality to find the $\vec u_2^{\,*}$ for which the first variation of \eqref{J2} is $0 \; \forall \, \delta \vec{u}_2^{\,*} \in \mathcal{C}^{0}$.  Using the G\^{a}teaux derivative and the definition of the functional derivative,
\newer{
\setlength{\arraycolsep}{0.0em}
\begin{eqnarray}
\label{deltaJ2}
\delta J_2 %
{ }=&{ }& \, \frac{d}{d \epsilon} \int_{t_0}^{t_f} l_2(t,\vec{u}_2^{\,*}(t)+\epsilon \, \vec \eta(t)) \,dt |_{\epsilon = 0}\nonumber\\
{ }=&{ }& \int_{t_0}^{t_f} \frac{\partial l_2(t,\vec{u}_2^{\,*}(t))}{\partial \vec{u}_2(t)} \, \vec \eta(t) \,dt \;{ }={ }\, 0  \;\;\;\; \forall \eta \text{\,,}%
\end{eqnarray}
\setlength{\arraycolsep}{5pt}
}\noindent
where $\epsilon$ is a scalar and $\epsilon \, \vec \eta = \delta \vec u_2^{\,*}$.  

\newer{A} generalization of the Fundamental Lemma of Variational Calculus \cite{naiduOptimalCont2002}, implies $\frac{\partial l_2(\cdot,\cdot)}{\partial \vec u_2} = \vec 0$ at the optimizer.  \newer{Solving
\begin{equation}
\label{partl2U2}
\frac{\partial l_2(\cdot,\cdot)}{\partial \vec u_2} = (\vec{\rho}^{\;T} \, \mat h(\vec x) \, [\vec u_2^{\,*} - \vec u_1] - \alpha_d) \vec{\rho}^{\;T} \, \mat h(\vec x) + \vec u_2^{\,*\,T} \mat R = \vec{0} \text{\,}%
\end{equation}
\noindent
in terms of $\vec u_2^{\,*}$ confirms the optimal schedule is \eqref{U2Opt}.}
\end{IEEEproof}

\newer{To summarize, in computing optimal actions, SAC calculates a schedule providing the value of the optimal action at every possible application time along the predicted trajectory.  These values optimize a local model of the change in cost relative to control duration at each application time.  The model is provided by the mode insertion gradient \eqref{dJdlambda}.  As a benefit of SAC, the schedule of optimal action values can be computed from a closed-form expression, \eqref{U2Opt}, of the nominal state and adjoint \eqref{rhodot} \emph{even for non-convex tracking costs \eqref{J}}.
}

\subsection{\newer{Deciding When to Act}}
\label{search}

\begin{figure}[t]
\centering
\includegraphics[height=1.1in]{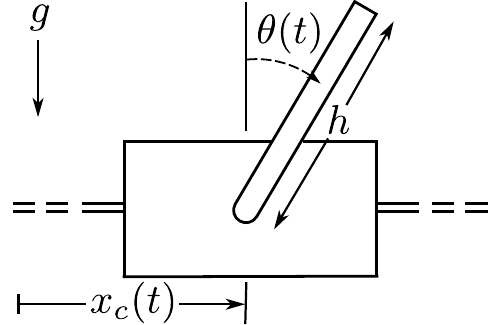}
\caption{Configuration variables for the cart-pendulum system.}
\label{fig:cartPendConfig}
\end{figure}

\begin{figure}[t]
\centering
\includegraphics[width=2.7in]{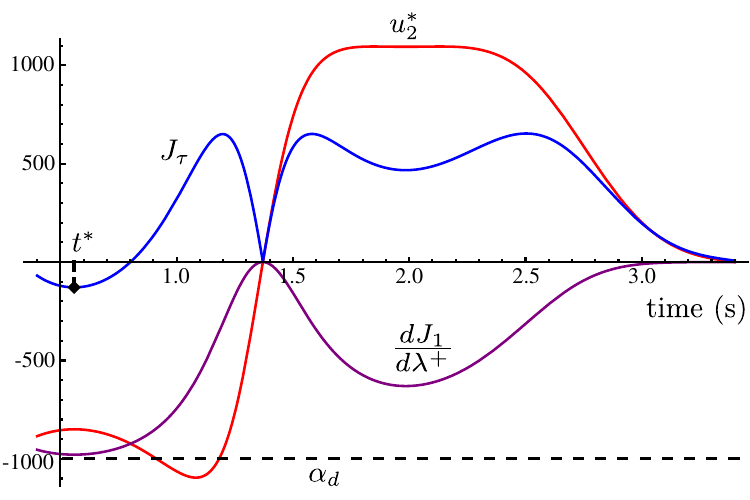}
\caption{A schedule of optimal actions, $\vec u_2^{\,*}$, is depicted (red curve) for a $T = 3 \, s$ predicted trajectory of the cart-pendulum system \eqref{fpend} starting at current time $t_0 = 0.4 \, s$.  These actions minimize acceleration in driving the mode insertion gradient toward $\alpha_d = -1,000$.  The (purple) mode insertion gradient curve approximates the change in cost \eqref{J} achievable by short application of $\vec u_2^{\,*}(t)$ at different times.  The objective, $J_{\tau}$, (blue curve) is minimized to find an optimal time, $t^*$, to act.  Waiting to act at $\tau=t^*$ rather than at $\tau=t_0$ \newer{(we assume $t_{calc}=0$),} SAC generates greater cost reduction using less effort.}
\label{fig:u2Search}
\end{figure}

\begin{newersection}
Assuming control calculations require some time, $t_{calc} < t_s$, SAC searches $\vec u_2^{\,*}$ for application times, $\tau \in (t_0+t_{calc},t_f)$,\footnote{SAC implements the previous action while current calculations complete.} that optimize an objective to find the most effective time to act over each predicted trajectory.
We use
\begin{equation}
\label{fcost}
J_{\tau}(t) = \norm{\vec u_2^{\,*}(t)} + \frac{dJ_1}{d \lambda^+}(t,u_2^{\,*}(t)) + (t-t_0)^{\, \beta} \text{\,,}
\end{equation}
to balance a trade-off between control efficiency and the cost of waiting, though there are many other choices of objective.\footnote{Implementation examples apply $\beta = 1.6$ as a balance between time and control effort in achieving tracking tasks, but any choice of $\beta > 0$ will work.}

Consider, for example, inverting a simple cart-pendulum with state, $\vec x = (\theta, \, \dot \theta, \, x_c, \, \dot x_c)$ as in Fig.~\ref{fig:cartPendConfig}, acceleration control, $\vec u = (a_c)$, and underactuated dynamics,
\begin{equation}
\label{fpend}
\vec f(\vec x, \vec u) = \left(\begin{IEEEeqnarraybox*}[][c]{,c,}
\dot \theta\\
\frac{g}{h} \, \sin(\theta) + \frac{a_c \, \cos(\theta)}{h}\\
\dot x_c\\
a_c%
\end{IEEEeqnarraybox*}\right)\text{\,,}
\end{equation}
with length, $h=2\,$m, and gravity, $g=9.81\,\frac{\text{m}}{\text{s}^2}$.  Fig.~\ref{fig:u2Search} shows a schedule, $\vec u_2^{\,*}$, computed for \eqref{fpend} starting at $t_0 = 0.4 \, s$ into an example closed-loop SAC trajectory.  
At every time, the action in $\vec u_2^{\,*}$ drives the mode insertion gradient (purple curve) toward $\alpha_d = -1,000$.  
The mode insertion gradient is $0$ at $t \approx 1.39\,$s when the pendulum is horizontal, i.e., $\theta = \frac{\pi}{2}\,$rad., since no action can push $\theta$ toward $\theta = 0$ at that time.
The mode insertion gradient also goes to $0$ toward the end of the horizon since no finite control action can improve \eqref{J} at the final time.  The curve of $J_{\tau}$ vs time (blue) indicates, to optimize the trade-off between wait time and effectiveness of the action (as in \eqref{fcost}), SAC should do nothing until optimal time $t^* \approx 0.57\,$s.\footnote{\newer{The next SAC synthesis cycle, i.e., the next sampling time, may begin (and conclude) before the action at $t^*$ is applied.  
In such cases, SAC often computes a similar $t^*$.  So, in this case, SAC would likely continue to wait until $t^* \approx 0.57\,$s to act.}} %
\end{newersection}

\subsection{\newer{Deciding How Long to Act}}
\label{duration}

Temporal continuity of $\vec \rho$, $\vec f_1$, $\vec f_2$ and $\vec u_1$ provided by Assump.~\ref{assump:dynam}-\ref{assump:control} ensures the mode insertion gradient %
is continuous with respect to duration around where $\lambda \rightarrow 0^+ \; \forall \; \tau \in (t_0,t_f)$.  Therefore, there exists a %
neighborhood, $V = \mathcal{N}(\lambda \rightarrow 0^+)$, where the sensitivity indicated by \eqref{dJdlambda} models the change in cost relative to application duration to first-order (see \cite{TimSuffDescentModeSched2016,MagnustInsertionGradientTransitionTime} \new{and the generalized derivation in Section~\ref{hybrid_sac}}). For finite durations, $\lambda \in V$, the change in cost \eqref{J} is locally modeled as
\begin{equation}
\label{DeltaJ}
\newer{\Delta J_1 \approx \frac{dJ_1}{d \lambda^+}(\tau,u_2^{\,*}(\tau)) \, \lambda \text{\,.}}
\end{equation}

As $\vec u_2^{\,*}(\tau)$ regulates $\newer{\frac{dJ_1}{d \lambda^+}(\tau,u_2^{\,*}(\tau))} \approx \alpha_d$, \eqref{DeltaJ} becomes $\Delta J_1 \approx \alpha_d \lambda$.  Thus the choice of $\lambda$ and $\alpha_d$ allows the control designer to specify the desired degree of change provided by actions, $\vec u_2^{\,*}(\tau)$. \new{\edit{We use a line search with a simple descent condition to find a $\lambda \in V$ that yields the desired change \cite{NocedalOptimization2006}.}}\footnote{\newer{Because the pair $(\alpha_d, \lambda)$ determines the change in cost each action can provide, it is worth noting that a sufficient decrease condition similar to the one proposed in \cite{TimSuffDescentModeSched2016} can be applied \new{to the choice of $\lambda$.}}}

\newer{Upon selection of the application duration, $\lambda$, the SAC action is fully specified and sent to the robot.  The process iterates and the next cycle begins when SAC incorporates new state feedback at the subsequent sampling time.}

\section{Special Properties of SAC Control}

In addition to providing a closed-form solution for the entire schedule of optimal actions \eqref{U2Opt}, SAC controls inherit powerful guarantees.  Appendix~\ref{app:A} includes derivations that show 1) the schedule \eqref{U2Opt} globally optimizes \eqref{u2sched}.  2) Around equilibria, SAC controls simplify to linear state feedback laws permitting local stability analysis and parameter selection, e.g., parameters of \eqref{J}, $\alpha_d$, or $T$. 3) Finally, actions computed from \eqref{U2Opt} can be saturated to satisfy min-max constraints using quadratic programming, by scaling the control vector, or by scaling components of the control vector.\footnote{Proofs are included for each with $\vec u_1 = \vec{0}$, as in all the examples in this paper. All examples enforce constraints using the component scaling approach.}

For an overview of the SAC approach outlining the calculations required for on-line synthesis of constrained optimal actions, selection of actuation times, and resolution of control durations, refer to Algorithm~\ref{sacAlgorithm}.

\section{Example Systems}
\label{examples}

The following section provides simulation examples that apply SAC on-line \new{in benchmark underactuated control tasks.}\footnote{\new{\edit{We also have trajectory tracking results, e.g, for differential drive robots, but cannot include them due to space constraints.}}}  Each example emphasizes a different performance-related aspect of SAC and results are compared to alternative methods.

\begin{figure}
\begin{minipage}[h!]{1.0\linewidth}
\begin{algorithm}[H]
  \caption{Sequential Action Control}\label{sacAlgorithm}
  \begin{algorithmic}[0]
    \State Initialize $\alpha_d$, minimum change in cost $\Delta J_{min}$, current time $t_{curr}$, default control duration $\Delta t_{init}$, nominal control $\vec u_1$, scale factor $\omega \in (0,1)$, prediction horizon $T$, sampling time $t_s$, the max time for iterative control calculations $t_{calc}$, and the max backtracking iterations $k_{max}$.
\While{$t_{curr} < \infty$}
    \State $(t_0,t_f) = (t_{curr}, t_{curr}+T)$
    \State \newer{Use feedback to initialize $x_{init} = x(t_0)$}
    \State Simulate $(\vec x,\vec \rho)$ from $\vec f_1$ for $t \in [t_0, t_f]$
    \State Compute initial cost $J_{1,init}$
    \State Specify $\alpha_d$ \footnotemark
    \State Compute $\vec u_2^{\,*}$ from $(\vec x,\vec \rho)$ using Theorem~\ref{uthrm}
    \State Specify / search for time, $\newer{\tau} > t_0 + t_{calc}$, to apply $\vec u_2^{\,*}$
    \State Saturate $\vec u_2^{\,*}(\newer{\tau})$
    \State Initialize $k = 0$, $J_{1,new} = \infty$
    \While{$J_{1,new}-J_{1,init}>\Delta J_{min}$ \textbf{and} $k \leq k_{max}$}
    \State $\newer{\lambda}$ = $\omega^{\, k} \Delta t_{init}$
    \State $(\tau_0,\tau_f) = (\newer{\tau} - \frac{\newer{\lambda}}{2}, \newer{\tau} + \frac{\newer{\lambda}}{2})$
    \State Re-simulate $\vec x$ applying $\vec f_2$ for $t \in [\tau_0, \tau_f]$
    \State Compute new cost $J_{1,new}$
    \State $k = k+1$
    \EndWhile
    \State $\vec u_1(t) = \vec u_2^{\,*}(\newer{\tau})$ $\forall t \in [\tau_0, \tau_f] \cap [t_0+\newer{t_{calc}}, t_0+t_s+t_{calc}]$
    \State \newer{Send updated $u_1$ to robot}
    \While {$t_{curr} < t_0+t_s$}
    \State Wait$(\,)$
    \EndWhile
    \EndWhile
  \end{algorithmic}
\end{algorithm}
\vspace{-8pt}
\captionof{algcap}{At sampling intervals SAC incorporates feedback and simulates the system with a nominal (typically null) control.  Optimal alternative actions are computed as a closed-form function of time.  A time is chosen to apply the control action.  A line search provides a duration that reduces cost.}
\vspace{-12pt}
\end{minipage}
\end{figure}

\begin{figure*}[t!]
\begin{subfigure}[b]{.5\textwidth}
\centering
\includegraphics[scale=.9]{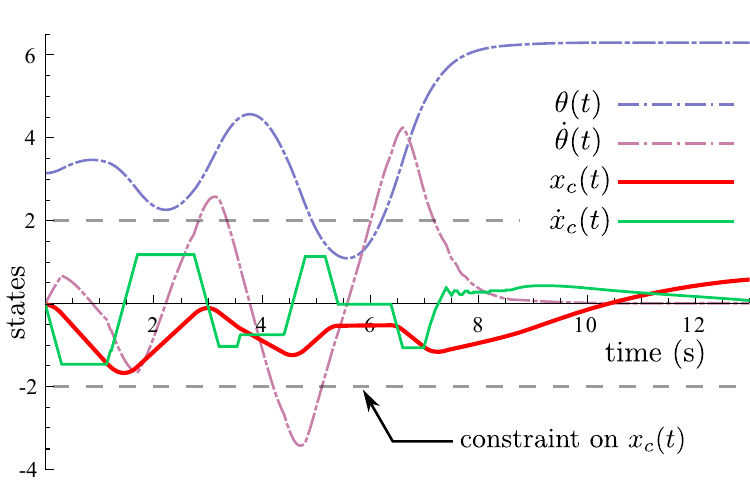}
\caption{ }
\label{fig:barrierStates}
\end{subfigure}
\hspace{0in}
\begin{subfigure}[b]{.5\textwidth}
\centering
\hspace{-3mm}
\includegraphics[scale=.9]{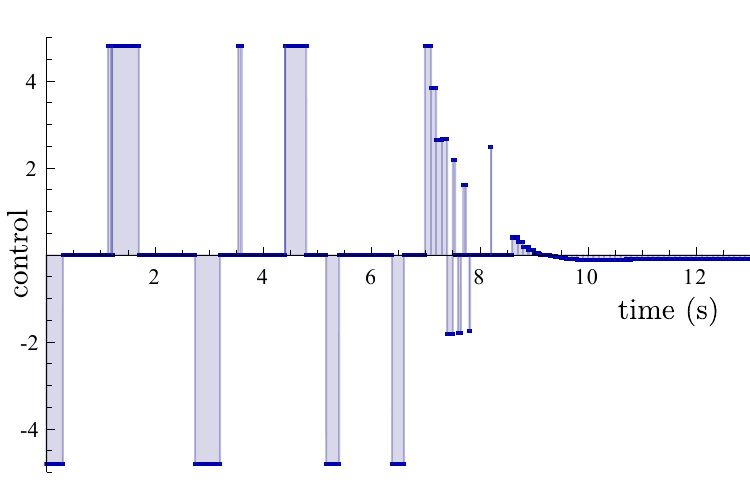}
\caption{ }
\label{fig:barrierControl}
\end{subfigure}
\caption{SAC inverts the cart-pendulum at a low sampling and control sequencing frequency of $10$ Hz (at equilibrium the dynamics correspond to a simple pendulum with natural frequency of $0.35$ Hz). This low-frequency control signal (Fig.~\ref{fig:barrierControl}) illustrates how individual actions are sequenced (especially apparent from $7$ to $10\,$s).  SAC maintains the cart in $[-2, 2\,]$ m during inversion.  Figure~\ref{fig:barrierControl} also shows SAC automatically develops an energy pumping strategy to invert the pendulum.}\label{fig:barrier}
\end{figure*}

\subsection{Cart-Pendulum}
\label{cartPend}

First, we present $3$ examples where SAC is applied to the nonlinear cart-pendulum \eqref{fpend} in simulated constrained \new{\edit{swing-up}}.  Performance of SAC is demonstrated using the cart-pendulum as it provides a well understood underactuated control problem that has long served as a benchmark for new control methodologies (see \cite{CartPendEnergy2000, CartPendLagran2005, CartPendSwingUpStable2000, CartPendRealSwingUp2006, CartPendGlobalStable2009, CartPendConstSwingUp2009}).

\footnotetext{\newer{In all examples, we choose to specify $\alpha_d$ as a feedback law, $\alpha_d = \gamma \, J_{1,init}, \gamma \in \reals{-}$.  We find $\gamma \in [-15, -1]$ works well.}\label{fn:alpha_d}}

\subsubsection{Low-Frequency Constrained Inversion}

This example uses SAC to invert the cart-pendulum \eqref{fpend} \new{with low frequency ($10$ Hz) feedback and control action sequencing to highlight the control synthesis process.}
\begin{figure*}[t]
\begin{subfigure}[b]{.5\textwidth}
\centering
\includegraphics[scale=.9]{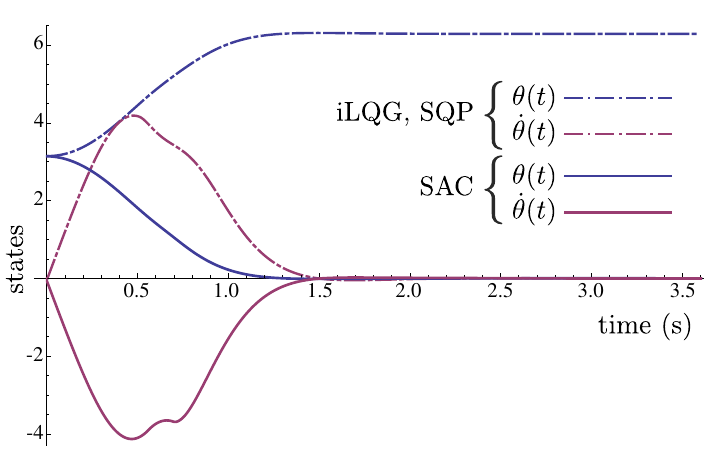}
\caption{ }
\label{fig:compStates}
\end{subfigure}
\hspace{0in}
\begin{subfigure}[b]{.5\textwidth}
\centering
\hspace{-3mm}
\includegraphics[scale=.9]{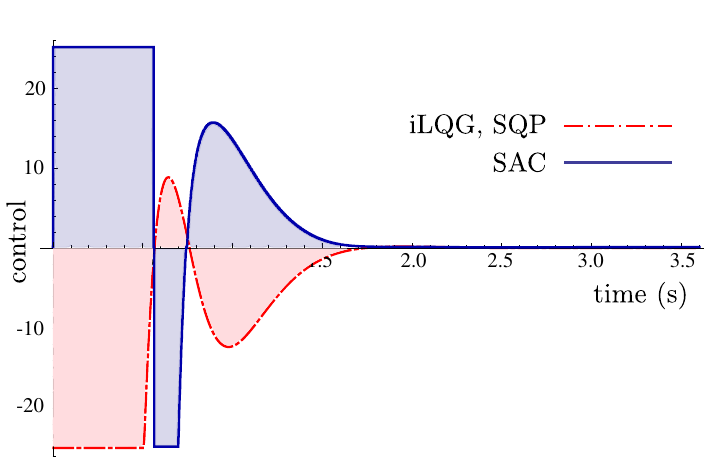}
\caption{ }
\label{fig:compControl}
\end{subfigure}
\caption{SAC can provide control solutions on-line and in closed-loop (these results reflect $1,000$ Hz feedback) that achieve performance comparable to or better than \new{solutions from nonlinear optimal control.  For the trajectory depicted, SAC achieves the same final cost of $J_{pend} \approx 2,215$ as SQP and iLQG.}}\label{fig:SACvsSQP}
\end{figure*}
\new{Control constraints, $\ddot x_c \in [-4.8, 4.8] \, \frac{m}{\text{s}^2}$, show SAC can find solutions that require multiple swings to invert.  We use a quadratic tracking cost \eqref{Jquad} with the state dependent weights, $\mat Q(\vec x(t)) = \mat{Diag}[\,200, 0, (x_c(t)/2)^8, 50\,]$.
to impose a barrier / penalty function (see \cite{BoydConvexOpt, OptControl2010}) that constrains the cart's state so $x_c \in [-2, 2\,]$.
Terminal and control costs in \eqref{l2} and \eqref{Jquad} are defined using $\mat{P_1} = \mat 0$, $\mat R = [0.3]$, and a horizon of $T = 1.5 \, $s.}\footnote{\new{All examples use wrapped angles $\in [-\pi, \pi)\, \text{rad}$.}}

\new{Results in Fig.~\ref{fig:barrier} correspond to an initial condition with the pendulum hanging at the stable equilibrium and zero initial velocity, $\vec x_{init} = ( \pi, 0 )$.  The red curve shows the penalty function successfully keeps the cart position within $[-2, 2\,]\,$m. The simulated trajectory is included in the video attachment.}

\subsubsection{High-Frequency Constrained Inversion}

In this example, SAC performs on-line swing-up and cart-pendulum inversion with high-frequency feedback ($1$ KHz).  
\new{To gauge the quality of the inversion strategy, we compare the on-line, closed-loop SAC control to off-line \newer{trajectory optimization}
using MATLAB's sequential quadratic programming (SQP) and iLQG implementations.
The SQP method is widely used and underlies the approach to optimization in \cite{robustSQP2008,snopt2002,SQPconstControl2004,ConstContrlSftwrSQP2010,NLPQLP2006,robustSQP2011}. 
The iLQG algorithm \cite{iLQGmpcTodorov2012,iLQGtodorov2005} is a state-of-the-art variant of differential dynamic programming (DDP).
While early versions did not accommodate control constraints, iLQG achieves a tenfold speed improvement over DDP in simulations \cite{iLQRtodorov2004} and has since been applied for real-time humanoid control \cite{iLQGmpcTodorov2012}.  This section compares to a recent variant that incorporates control constraints through a new active-set method \cite{iLQGcontConst2014}.  We use a publicly available MATLAB iLQG implementation developed by its authors.\footnote{\new{Available at \url{http://www.mathworks.com/matlabcentral/fileexchange/52069-ilqg-ddp-trajectory-optimization}}.}}

To highlight the sensitivity of optimal control, i.e., iLQG and SQP, to local minima even on simple nonlinear problems (and to speed SQP computations), this example uses a low-dimensional cart-pendulum model.  The simplified model leaves the cart position and velocity unconstrained and ignores their error weights, such that dynamics are represented by the first two components of \eqref{fpend}.  
In this case, the goal is to compute controls that minimize a norm on the cart's acceleration while driving the pendulum angle toward the origin (inverted equilibrium).
We compare performance of the trajectories produced by each algorithm over a fixed time horizon, $T_{opt}$, based on the objective,
\begin{equation}
\label{Jsqp}
J_{pend} = \frac{1}{2} \int_{0}^{T_{opt}} \norm{\vec x(t) - \vec x_d(t)}_{\mat Q}^2 + \norm{\vec u(t)}_{\mat R}^2 \,dt \text{\,,}
\end{equation}
with $\mat Q = \mat{Diag}[1000, \, 10]$ and $\mat R = [0.3]$.
All algorithms are constrained to provide controls $\ddot x_c \leq \abs{25} \, \frac{m}{\text{s}^2}$.

\begin{table}
\centering
\new{
\begin{tabular}{cc|*{6}{c|}}  %
\cline{3-8}
& & \multicolumn{2}{c}{$T_{opt}=4\,$s} & \multicolumn{2}{|c}{$T_{opt}=5\,$s} & \multicolumn{2}{|c|}{$T_{opt}=6\,$s}\\ \cline{3-8}
dt &   & min. & iters & min. & iters & min. & iters \\ \hline
\multirow{2}{*}{$.01\,$s} & SQP & 13 & 1,234& \textcolor{gray}{22}& \textcolor{gray}{1052}& \textcolor{gray}{46}& \textcolor{gray}{1,346}\\
& iLQG & 2& 737& 9& 2,427& 13& 3,108\\ \hline
\multirow{2}{*}{$.005\,$s} & SQP & 169& 2,465& \textcolor{gray}{32}& \textcolor{gray}{201}& \textcolor{gray}{105}& \textcolor{gray}{251} \\
& iLQG & 5& 908& 56& 8,052& \textcolor{gray}{5}& \textcolor{gray}{622}\\ \hline
\multirow{2}{*}{$.003\,$s} & SQP & 689 & 2,225 & \textcolor{gray}{817}& \textcolor{gray}{853}& \textcolor{gray}{1,286} & \textcolor{gray}{933}\\
& iLQG & 9& 1,007& 28& 2,423& \textcolor{gray}{9}& \textcolor{gray}{688}\\ \hline
\end{tabular}
\caption{
SQP versus iLQG for swing-up of the cart-pendulum under varying optimization horizon, $T_{opt}$, and discretization, dt. All solutions converge to the same optimizer with $J_{pend} \approx 2,215$, except the gray results, which converged to low performance local minima. For each parameter combination, columns indicate the number of iterations (iters) and time in minutes (min.) for convergence. 
}\label{tab:SQPviLQG}
}
\end{table}

Both SQP and iLQG directly optimize a discretized version of \eqref{Jsqp} to derive their optimal trajectories.  For comparison, results are provided for different choices of discretization, dt, and optimization horizons, $T_{opt}$.\footnote{\new{Horizons are based on the assumed time for pendulum inversion, and discretizations on assumed frequency requirements and linearization accuracy.}}
In contrast, SAC computes a trajectory of duration $T_{opt}$ by deriving actions from a receding state tracking cost \eqref{Jquad} (see Appendix~\ref{app:guarantees}) with quadratic state norms similar to the one in \eqref{Jsqp}.
Although SAC runs at $1$ KHz, optimal control results are limited to dt$\,\geq 0.003\,$s, as SQP computations become infeasible  
and consume all computational resources below this.\footnote{All results were obtained on the same laptop with Intel$^{\circledR}$ Core\textsuperscript{TM} i7-4702HQ CPU @ 2.20GHz $\times$ 8 and 16GB RAM.}
Table~\ref{tab:SQPviLQG} provides the time and number of optimization iterations required for each parameter combination.

\new{The parameter combinations in Table~\ref{tab:SQPviLQG} that do not correspond to gray data converged to the same (best case) optimal trajectory, %
which inverts the pendulum in $< 2\,$s with $J_{pend} \approx 2,215$.
\footnote{The cost of the optimal solution is the same when measured for horizons $T_{opt}=4-6\,$s since the incremental cost in \eqref{Jsqp} is negligible after inversion at $t \approx 2\,$s.}
Gray data indicate convergence to an alternate local minima with significantly worse cost. In all cases with $T_{opt} \neq 4\,$s, SQP converges to local minima with costs $J_{pend} \approx 3,981 - 6,189$.  While iLQG tends to be less sensitive to local minima, it converges to the worst local minima with $J_{pend} \approx 9,960$ for both finer discretizations when $T_{opt} = 6\,$s.}

Since varying $T_{opt}$ has no affect on SAC control synthesis (other than to specify the duration of the resulting trajectory), 
SAC control simulations included a
variety of additional parameter combinations including receding horizons from $T = 0.15\,\text{s} - 3 \, $s and different synthesis frequencies.  These solutions yield costs %
ranging from $J_{pend} = 2,215 - 2,660$, with the majority of solutions close or equal to $J_{pend} = 2,215$.  
The SAC solution depicted in Fig.~\ref{fig:SACvsSQP} achieves the best case cost of $J_{pend} = 2,215$  
from receding horizons of $T = 0.28 \,$s, with parameters $\mat Q = \mat 0$ and $\mat{P_1} = \mat{Diag}[500, \, 0]$ in \eqref{Jquad}, and with $\mat R = [0.3]$.
\newer{SAC's on-line controls} perform constrained inversion as well as the best solutions from offline optimal control.
Also, local minima significantly affect SQP and iLQG, while SAC tends to be less sensitive.

\new{Considering the simplicity of this nonlinear example, it is noteworthy that both optimal control algorithms require significant time to converge.
While iLQG ranges from minutes to an hour, 
with a discretization $3\times$ as coarse as SAC, SQP requires $\approx 12$ hours to compute the single, open-loop \newer{trajectory} in Fig.~\ref{fig:SACvsSQP} using 4 CPU cores.  Our C++ implementation of SAC obtains a solution equivalent to the best results on-line, with $1$ KHz feedback, in less than $\frac{1}{2}\,$s using 1 CPU core.\footnote{\new{As the MATLAB SQP and iLQG implementations utilize compiled and parallelized libraries, it is unclear how to provide a side-by-side comparison to the timing results in Table~\ref{tab:SQPviLQG}.  To illustrate that SAC is still fast in slower, interpreted code, we also implemented SAC in Mathematica.  Computations require $5 - 35\,$s and are linear w.r.t. to horizon, $T$, and discretization, $t_s$.}}
Computing optimal actions in closed-form, SAC achieves dramatic gains and avoids the iterative optimization process, which requires thousands of variables and constraints in SQP / iLQG.}

\new{
Finally, we emphasize the closed-loop nature of SAC compared to SQP, which provides an open-loop trajectory, and iLQG, which yields an affine controller with both feedforward and feedback components.  
As the affine controller from iLQG is only valid \newer{near} the optimal solution (SAC provides feedback from arbitrary states), SQP or iLQG must be applied in receding horizon for feedback comparable to SAC.  
For improved speed, \cite{iLQGcontConst2014} recommends a receding horizon implementation using suboptimal \newer{solutions from a fixed number (one) of iterations.}
However, in this simple nonlinear example, SQP / iLQG trajectories only resemble the final solution a few iterations before convergence.
\newer{Hence, receding horizon implementations would likely result in poor local solutions.}}%

\subsubsection{Sensitivity to Initial Conditions}

Using a horizon of $T = 1.2 \,$s, SAC was applied to invert the same, reduced cart-pendulum system from a variety of initial conditions.  Simulations used the quadratic tracking cost \eqref{Jquad} and weight matrices from \eqref{Jsqp}. A total of $20$ initial conditions for $\theta(t)$, uniformly sampled over $[0, 2 \, \pi)\,$rad, were paired with initial angular velocities at $37$ points uniformly sampled over $[0, 4 \, \pi]\, \frac{\text{rad}}{\text{s}}$.

To gauge performance, a $10 \,$s closed-loop trajectory was constructed from each of the $740$ sampled initial conditions, and the state at the final time $x(10 \, \text{s})$ measured.  If the final state was within $0.001$ rad of the inverted position and the absolute value of angular velocity was $< 0.001 \, \frac{\text{rad}}{\text{s}}$, the trajectory was judged to have successfully converged to the inverted equilibrium.  Tests confirmed the SAC algorithm was able to successfully invert the pendulum within $10 \,$s from all initial conditions.  The average computation time was $\approx 1\,$s for each $10\,$s trajectory on the test laptop.

\begin{figure}[t]
\centering
\includegraphics[width=2.5in]{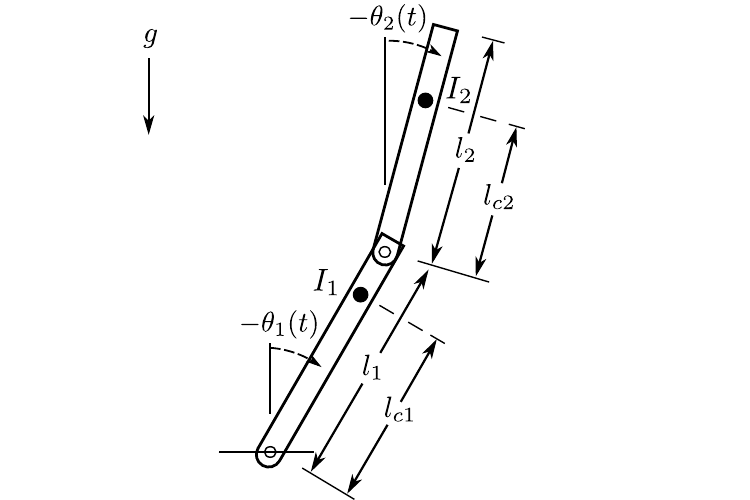}
\caption{Configuration of the acrobot and pendubot systems.  
}
\label{fig:AcroPendConfig}
\end{figure}

\subsection{Pendubot and Acrobot}
\label{pendubot}

This section applies SAC for swing-up control of the pendubot \cite{Pend2009,PendExper2006,SpongPend1995} and acrobot \cite{SpongAcro1995,AcroSwingUpAnalysis2007,AcroEnergySwingUp2012}.  The pendubot is a two-link pendulum with an input torque that can be applied about the joint constraining the first (base) link.  The acrobot is identical except the input torque acts about the \new{second joint.}  The nonlinear dynamics and pendubot model parameters match those from simulations in \cite{Pend2009} and experiments in \cite{PendExper2006}. The acrobot model parameters and dynamics are from simulations in \cite{AcroSwingUpAnalysis2007} and in seminal work \cite{SpongAcro1995}.  Figure~\ref{fig:AcroPendConfig} depicts the configuration variables and the model parameters are below.  Each system's state vector is $\vec x = (\theta_1, \, \dot \theta_1, \, \theta_2, \, \dot \theta_2)$ with the relevant joint torque control, $\vec u = (\tau)$. 

\vspace{0.5mm}
\begin{center}
\begin{tabular}{lll}
\textbf{pendubot:} & m$_{\text{1}}$ = 1.0367 kg & m$_{\text{2}}$ = 0.5549 kg\\
 & l$_{\text{1}}$ = 0.1508 m & l$_{\text{2}}$ = 0.2667 m\\
 & l$_{\text{c1}}$ = 0.1206 m & l$_{\text{c2}}$ = 0.1135 m\\
 & I$_{\text{1}}$ = 0.0031 kg m$^{\text{2}}$ & I$_{\text{2}}$ = 0.0035 kg m$^{\text{2}}$\\
 &  & \\
\textbf{acrobot:} & m$_{\text{1}}$ = 1 kg & m$_{\text{2}}$ = 1 kg\\
 & l$_{\text{1}}$ = 1 m & l$_{\text{2}}$ = 2 m\\
 & l$_{\text{c1}}$ = 0.5 m & l$_{\text{c2}}$ = 1 m\\
 & I$_{\text{1}}$ = 0.083 kg m$^{\text{2}}$ & I$_{\text{2}}$ = 0.33 kg m$^{\text{2}}$\\
\end{tabular}
\end{center}
\vspace{1mm}

Due to their underactuated dynamics and many local minima, the pendubot and acrobot provide challenging test systems for control.  As a popular approach, researchers often apply energy based methods for swing-up control and switch to LQR controllers for stabilization in the vicinity of the inverted equilibrium (see \cite{Pend2009,SpongEnergyPend2000,UnifUnder2linkManipControl2009,SpongAcro1995,SpongPend1995,AcroSwingUpAnalysis2007,AcroEnergySwingUp2012}).  %
\new{We also use LQR controllers to stabilize the systems once near the inverted equilibrium.}  However, the results here show \newer{SAC} can swing-up both systems without \newer{special} energy optimizing methods.  The algorithm utilizes the quadratic state error based cost functional \eqref{Jquad}, without modification.

While the pendubot simulations in \cite{Pend2009} require control torques up to a magnitude of $\SI{15}{\newton\meter}$ for inversion, the experimental results in \cite{PendExper2006} perform inversion with motor torques restricted to $\pm\SI{7}{\newton\meter}$.  Hence, the pendubot inputs are constrained to $\tau \in [-7, 7]\,\SI{}{\newton\meter}$.  The acrobot torques are constrained with $\tau \in [-15, 15]\,\SI{}{\newton\meter}$ to invert the system using less than the $\SI{20}{\newton\meter}$ required in \cite{AcroSwingUpAnalysis2007}.

\new{Example simulations initialize} each system at the downward, stable equilibrium and the desired position is the \newer{fully inverted equilibrium.}
\new{Results} are based on a feedback sampling rate of $200\,$Hz for the pendubot with $\mat Q = \mat{Diag}[100, \, 0.0001, \, 200, \, 0.0001]$, $\mat{P_1} = \mat{0}$, and $\mat R = [0.1]$ and $400\,$Hz for the acrobot with $\mat Q = \mat{Diag}[1,000, \, 0, \, 250, \, 0]$, $\mat{P_1} = \mat{Diag}[100, \, 0, \, 100, \, 0]$, and $\mat R = [0.1]$.  The LQR controllers derived offline for final stabilization,
$\vec K_{lqr} = (-0.23, \, -1.74, \, -28.99, \, -3.86 \, )$
and
$\vec K_{lqr} = (-142.73, \, -54.27, \, -95.23, \, -48.42 \, ) \text{\,,}$
were calculated about the inverted equilibrium to stabilize the pendubot and acrobot\newer{, respectively.}  %
\new{We} selected  $|\theta_{1,2}| \leq 0.05$ as the switching condition for pendubot stabilization.\footnote{More formally, a supervisory controller can switch between swing-up and stabilizing based on the stabilizing region of attraction \cite{SpongUnderactuated1998, SpongReacWheelPend2001}.}  %
\new{The acrobot switched} %
once all its configuration variables were $\leq \abs{0.25}$.

\begin{figure}[t]
\centering
\includegraphics[width=2.8in]{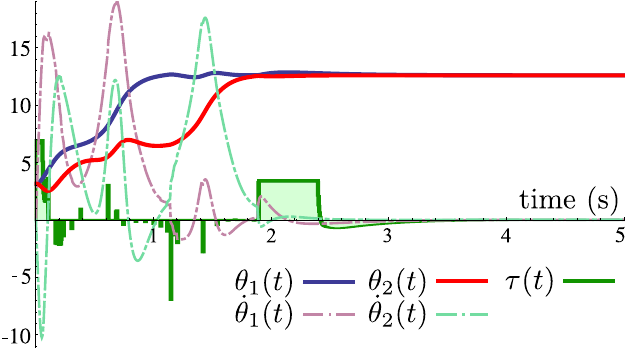}
\caption{\new{SAC swings up the pendubot close enough for final stabilization by the LQR controller.}  The LQR controller takes effect at $t=1.89$ s.  The algorithm inverts the system using less peak control effort and in less time than existing methods from literature with the same parameters.}
\label{fig:pendubot}
\end{figure}

\new{Figure~\ref{fig:pendubot} shows the pendubot trajectory (the acrobot and pendubot solutions are in video attachment).} In both cases, SAC swings each system close enough for successful stabilization.  %
\newer{SAC} inverts the pendubot using the same peak effort as in experiments from \cite{PendExper2006} and less than half that from simulations in \cite{Pend2009}.  Also, SAC requires only $3\,$s to invert, while simulations in \cite{Pend2009} needed $\approx 4\,$s.  \newer{Where %
\cite{Pend2009} %
switches between} separately derived controllers for pumping energy into, out of, and inverting the system before final stabilization, \emph{SAC performs all these tasks without any change in parameters} and with the simple state tracking norm in \eqref{Jquad}.  \new{In the case of the} acrobot, SAC inverts the system with the desired peak torque magnitude of $\SI{15}{\newton\meter}$ ($\frac{3}{4}$ the torque required in simulations from \cite{AcroSwingUpAnalysis2007}).  These closed-loop results were computed on-line and required only 1.23 and $4.7\,$s to compute $20\,$s trajectories for the pendubot and acrobot systems, respectively.

\new{To} invert the pendubot and acrobot in minimal time and under the tight input constraints, the two most important parameters for tuning are the horizon length, $T$, and the desired change in cost due to each control actuation, $\alpha_d$.
All examples specify $\alpha_d$ iteratively based on the current initial trajectory cost under the nominal (null) control as  $\alpha_d = \gamma \, J_{1,init}$.
Generally, because of the speed of SAC computations, good parameters values can be found relatively quickly using sampling.
These pendubot and acrobot results use $\gamma = -15$ and similar horizons of $T = 0.5\,$s and $T = 0.6\,$s, respectively.

\new{\edit{As mentioned earlier, optimal controllers typically use energy metrics for swing-up of the pendubot and acrobot, as simple state-tracking objectives yield local minima and convergence to undesirable solutions. 
It is noteworthy that SAC is able to invert both systems on-line and at high frequency considering optimal controllers (SQP/iLQG) generally fail under the same objective \eqref{Jquad}.}}

\begin{newsection}

\section*{Part II: Extension to Hybrid Impulsive Systems}

Part II of this paper extends SAC to systems with hybrid impulsive dynamics.
These systems model a more general class of robotics problems in locomotion and manipulation, which involve contact and impacts.
Such systems are challenging in optimal control and require specialized treatment and optimality conditions 
\cite{BouncingiLQGmpcTodorov2011,sussHybridPMP1999,iLQGmpcTodorov2012}.  \newer{By planning each control action in a neighborhood of 0 duration, SAC avoids these issues and does not need to optimize control curves over discontinuous segments of trajectory.} 
\section{\newer{Control Synthesis for Hybrid Systems}}
\label{hybrid_sac}

The SAC algorithm introduced in Section~\ref{control} is limited to differentiable nonlinear systems because the mode insertion gradient \eqref{dJdlambda} is subject to Assump.~\ref{assump:dynam}.
\newer{Rather than rely on \eqref{dJdlambda}, this section directly develops}
a first-order approximation of the variation in state and cost due to \newer{the perturbation in nominal control generated by each SAC action}.
\newer{We show the change in cost due to short SAC actions corresponds to the same mode insertion gradient formula \eqref{dJdlambda}, but in terms of an adjoint variable derived for hybrid impulsive systems.}
As a result (and a benefit of SAC), the SAC process described in Algorithm~\ref{sacAlgorithm} remains unchanged for hybrid impulsive systems.

Section~\ref{sec:hyb_examps} demonstrates the hybrid calculations on a $1D$ system and then illustrates SAC in simulated on-line control of a bouncing ball.  
The section concludes with %
\newer{the spring-loaded inverted pendulum (SLIP) example from the introduction.}

\subsection{\newer{Prediction}}

\newer{As in Part I, SAC predicts the nominal motion of hybrid robotic systems and computes actions that improve trajectory cost over (receding) horizons.  However,
this section introduces new notation more appropriate for hybrid impulsive systems. Specifically,}
the classes of hybrid systems considered here are similar to those in \cite{sussHybridPMP1999} and are defined such that:%
\footnote{\new{We assume actions are not applied at switching times, exclude Zeno behavior, and allow only a single element of $\Phi$ to be active to exclude simultaneous events and potentially indeterminate behavior.
These (and continuity) assumptions guarantee a local neighborhood exists such that perturbed system trajectories evolve through the same nominal location sequence (as in \cite{sussHybridPMP1999}).}}

\begin{enumerate}
\item $\mathcal{Q}$ is a finite set of \emph{locations}.%
\item $\mathcal{M} = \{ \mathcal{M}_q \subseteq \reals{n_q} \}_{q \in \mathcal{Q}}$ is a family of \emph{state space} manifolds indexed by $q$.
\item $U = \{ U_q \subset \reals{m_q} \}_{q \in \mathcal{Q}}$ is a family of \emph{control spaces}.
\item $f = \{f_q \in \mathcal{C}( \mathcal{M}_q \times U_q , T\mathcal{M}_q )\}_{q \in \mathcal{Q}}$ is a family of maps to the tangent bundle, $T\mathcal{M}_q$.  The maps $f_q(x,u) \in T_x\mathcal{M}_q$ are the \emph{dynamics} at $q$.
\item $\mathcal{U} = \{ \mathcal{U}_q \subseteq \mathcal{L}(\subset \reals{}, U_q) \}_{q \in \mathcal{Q}}$ is a family of sets of admissible control mappings.
\item $\mathcal{I} = \{ \mathcal{I}_q \subset \reals{+} \}_{q \in \mathcal{Q}}$ is a family of consecutive subintervals corresponding to the time spent at each location $q$.
\item The series of \emph{guards}, $\Phi = \{\Phi_{q,q'} \in \mathcal{C}^1(\mathcal{M}_{q},\reals{}) : (q, q') \in \mathcal{Q}\}$, indicates transitions between locations $q$ and $q'$ when $\Phi_{q,q'}(x) = 0$.
The state transitions according to a series of corresponding \emph{reset maps}, $\Omega = \{\Omega_{q,q'} \in \mathcal{C}^1(\mathcal{M}_{q},\mathcal{M}_{q'}) : (q,q') \in \mathcal{Q}\}$.
\end{enumerate}

\newer{For clarity, we avoid using numerical subscripts for the nominal control, $u_1$.  Instead, SAC predicts nominal motions assuming a (possibly null) nominal control, \new{$u_{n,q} \in \mathcal{U}_q$, is defined for every location, $\forall q\in \mathcal{Q}$.  So, for instance, in the SLIP example, one nominal control is defined in stance with another, possibly identical, control in flight.  Note that as a hybrid robotic system applies controls, it evolves through an ordered sequence of locations, $(q_1, \dots, q_r) : r \in \nats{}$, e.g., from flight, to stance, to flight again, for a hopping SLIP.}} 

\newer{With the initial location as $q_1$, state $x(t_0) = x_{init} \in \mathcal{M}_{q_1}$ and the collection $\{f,\Phi,\Omega\}$, SAC's prediction phase simulates 
\begin{equation}
\label{hyb_xdot}
\dot x_{n,q_i} = f_{q_i}(x_{n,q_i},u_{n,q_i}) : t \in \mathcal{I}_{q_i}, q_i \in \mathcal{Q} \text{\,,}
\end{equation}
starting with $i=1$, to obtain the nominal state.  Guards indicate when a transition should occur, i.e., they specify the end of each interval $\mathcal{I}_{q_i}$, and the next location, $q_{i+1}$, based on which guard becomes $0$. Reset maps define the initial condition in $q_{i+1}$ as $\{x_{n,q_{i+1}}(t_i^+) = \Omega_{q_i,q_{i+1}}(x_{n,q_i}(t_i^-)) : t_i^- \triangleq \sup \mathcal{I}_{q_{i}}, t_i^+ \triangleq \inf \mathcal{I}_{q_{i+1}}\}$.  Through this process, the prediction phase defines the nominal location sequence, $(q_1, \dots, q_r)$, intervals, $\mathcal{I}$, and the resulting nominal trajectory,}
\[
\newer{(x_n(t), u_n(t)) \triangleq (x_{n,q_i}(t), u_{n,q_i}(t)) : i \in \{1,\dots,r\}, t \in \mathcal{I}_{q_i} \text{\,.}}
\]

\newer{As before, SAC's prediction phase concludes after computing the performance of the nominal trajectory.  However, in this hybrid case we use an objective,
\begin{equation}
\label{hyb_J}
J = \int_{t_0}^{t_f} l(x(t),u(t)) dt + m(x(t_f)) \text{\,,}
\end{equation}
with incremental and terminal costs defined in each location, $\{ l_{q_i} \in \mathcal{C}^1(\mathcal{M}_{q_i} \times U_{q_i},\reals{}) \}_{q_i \in \mathcal{Q}}$ and $\{ m_{q_i} \in \mathcal{C}^1(\mathcal{M}_{q_i},\reals{}) \}_{q_i \in \mathcal{Q}}$, such that $l = l_{q_i} : t \in \mathcal{I}_{q_i}$ and $m = m_{q_i} : t \in \mathcal{I}_{q_i}$. Also, \eqref{hyb_J} is more general than \eqref{J}, 
as it may depend on a control. Given $(x_n,\mathcal{I},q_1, \dots, q_r)$ resulting from nominal control, $u_n$, \eqref{hyb_J} can be evaluated along the hybrid trajectory.}
\subsection{\newer{Computing Optimal Actions}}

\newer{Recall that each cycle of SAC seeks an action that improves nominal trajectory performance.  %
This section defines the perturbed signal from an arbitrary SAC action of value $w$ as}
\begin{displaymath}
   u_w \triangleq \left\{
     \begin{array}{lr}
       u_n & : t \notin [\tau-\epsilon a, \tau]\\
       w & : t \in [\tau-\epsilon a, \tau]
     \end{array}
   \right.\text{\,,}
\end{displaymath} 
\newer{assuming a short duration,} $\lambda = \epsilon a$.  \newer{In this case,} the magnitude of $\lambda$ is specified as $\epsilon \in \reals{+}$ and the direction by an arbitrary positive scalar, $a \in \reals{+}$.  Because the perturbed system will eventually be evaluated as $\lambda \rightarrow 0^+$, assume the perturbation occurs \newer{when the nominal state, $x_n$, is in the arbitrary location $q_i$} so that $[\tau-\epsilon a, \tau] \subseteq \mathcal{I}_{q_i}$.\footnote{\newer{In the limit as $\lambda \rightarrow 0^+$, the SAC action is a needle perturbation \cite{PontryaginOptProcesses1962}.}}  Figure~\ref{fig:variations} depicts the perturbed control and the corresponding perturbed state.

\newer{To derive actions that maximally improve the nominal trajectory, Sec.~\ref{sec:actions} used the mode insertion gradient \eqref{dJdlambda} to model the change in cost \eqref{J} relative to control duration. To accommodate the discontinuous trajectories of hybrid robotic systems, this section derives a model of the change in nominal cost \eqref{hyb_J} resulting from the perturbed, $u_w$, by first modeling the effect of the control perturbation on state trajectory.
To these ends, we define the first-order perturbed state model,}\footnote{\new{The litte-o notation, $o(\epsilon)$, indicates terms that are higher than first order in $\epsilon$.  These terms go to zero faster than first-order terms in \eqref{hyb_xw} as $\epsilon \rightarrow 0$.}}
\begin{equation}
\label{hyb_xw}
x_w(t,\epsilon) \triangleq x_n(t) + \epsilon \Psi(t) + o(\epsilon)\text{\,.}
\end{equation}
\newer{The $\Psi(t)$ term is known as the \emph{variational equation} \cite{liberzonOptControl2012,PontryaginOptProcesses1962}.  It is the direction of the state variation at time $t$ and $\epsilon$ is the magnitude.  The following proposition provides formulas to compute the variational equation along hybrid impulsive trajectories.  The derivation is in Appendix~\ref{sec:variational_eq}.}
\begin{figure}[t]
  \centering
  \includegraphics[width=2.8in]{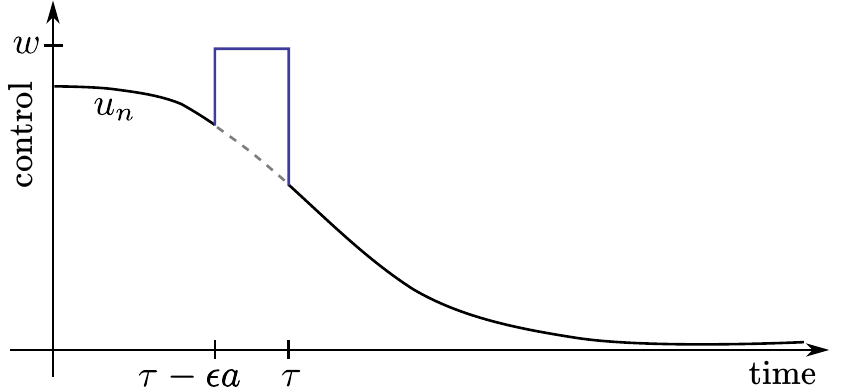}
  \includegraphics[width=2.8in]{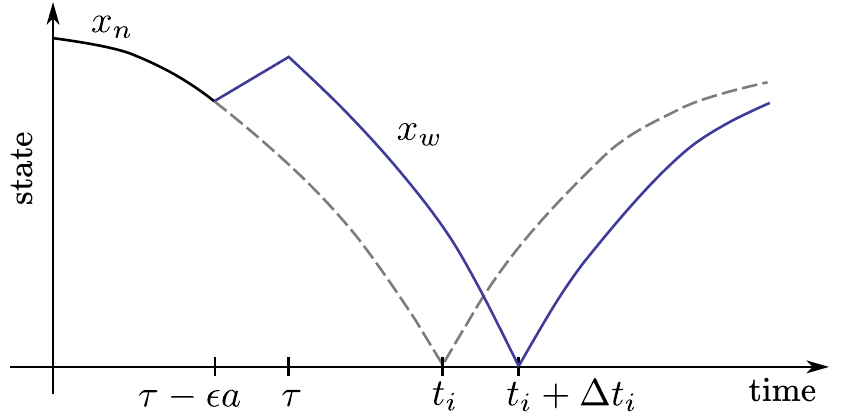}
  \caption{A perturbed control (top) and the corresponding state variation (bottom) for a hybrid system.  The nominal system switches locations at time $t_i$ and the perturbed system switches at time $t_i + \Delta t$.  Taken in the limit as $\epsilon a \rightarrow 0^+$, the control perturbation is a needle perturbation, which is equivalent to an infinitesimal duration action in SAC.}
\label{fig:variations}
\end{figure}
\newer{ 
\begin{proposition}
\label{prop:variational_eq}
Assume the state, $x_n$, of a hybrid system evolves from location $q_i \in \mathcal{Q}$ to $q_{i+1} \in \mathcal{Q}$ with the transition time, $t=t_i$.  If a control perturbation occurs at $t=\tau < t_i$, as in Fig.~\ref{fig:variations}, state variations propagate according to
\begin{equation}
\label{hyb_psi_dot_complex}
   \Psi \triangleq \left\{
     \begin{array}{ll}
       \bigg ( f_{q_i}(x_n(\tau),w) - f_{q_i}(x_n(\tau),u_n(\tau)) \bigg ) a & \hspace{-4pt}: t = \tau \in \mathcal{I}_{q_i}\\
       \dot \Psi = A_{q_i} \Psi & \hspace{-16pt} : t \in (\tau, t_i^-]\\
       \Psi(t_i^+) = \Pi_{q_i,q_{i+1}} \Psi(t_i^-) & \hspace{-16pt} : t = t_i^+\\
       \dot \Psi = A_{q_{i+1}} \Psi & \hspace{-16pt} : t \in (t_i^+, t_{i+1}^-]
     \end{array}
   \right.\hspace{-12pt}\text{,}
\end{equation}
with the linear variational reset map,
\setlength{\arraycolsep}{0.0em}
\begin{eqnarray}
\label{hyb_pi}
\Pi_{q_i,q_{i+1}} { }\triangleq{ } &&D_x \Omega_{q_i,q_{i+1}}(x_n(t_i^-)) \bigg [ I -f_{q_i}^-\nonumber\\
{}&&\frac{D_x\Phi_{q_i,q_{i+1}}(x_n(t_i^-))}{D_x\Phi_{q_i,q_{i+1}}(x_n(t_i^-))f_{q_i}^-} \bigg ] + f_{q_{i+1}}^+ \nonumber\\
{ }&&\frac{D_x\Phi_{q_i,q_{i+1}}(x_n(t_i^-))}{D_x\Phi_{q_i,q_{i+1}}(x_n(t_i^-))f_{q_i}^-} \text{\,,}%
\end{eqnarray}
$f_{q_i}(x_n(t_i^-),u_n(t_i^-)) \triangleq f_{q_i}^-$, $f_{q_{i+1}}(x_n(t_i^+),u_n(t_i^+)) \triangleq f_{q_i}^+$, and $A_{q_i}(t) \triangleq D_x f_{q_i}(x_n(t),u_n(t)) : t \in \mathcal{I}_{q_i}$ is the linearization about the (known) nominal state trajectory at $q_i$.
\end{proposition}
}\noindent

\newer{Note that if a nominal trajectory evolves through more than two locations,} each transition requires reset of $\Psi$ at transition times according to \eqref{hyb_pi}.  Variations continue according to the dynamics linearized about the nominal trajectory.  Repeating computations in rows $2-4$ of \eqref{hyb_psi_dot_complex} between consecutive locations, variations can be propagated to $t = t_f$.

\newer{With Prop.~\ref{prop:variational_eq} to compute the perturbed state \eqref{hyb_xw}, the following section derives the cost variation resulting from the perturbed control, $u_w$.  We will show the formula is a generalization of the mode insertion gradient \eqref{dJdlambda} that applies to a larger class of hybrid and impulsive systems.}

\subsubsection{\newer{Modeling the Cost Variation}}
\label{sec:sens_to_variat}

\newer{To first-order, the perturbed cost can be modeled as,
\begin{equation}
\label{hyb_Jw}
J_w(x_n, u_n,\epsilon) \triangleq J |_{(x_n, u_n)} + \epsilon \nu(t_f) + o(\epsilon) \text{\,,}
\end{equation}
where $\nu(t_f)$ is the direction of variation in the cost function and $\epsilon$ is the magnitude.
To simplify derivation of $\nu(t_f)$, we translate 
the hybrid system to Mayer form by appending the incremental costs, $l_{q_i}$, to the dynamics vectors, $f_{q_i}$, in each location. Objects with a bar refer to appended versions of hybrid system such that $\bar f_{q_i} = [\, l_{q_i},\, f_{q_i}^{\;T}\, ]^T$ 
and }
\setlength{\arraycolsep}{5pt}
\begin{displaymath}
  \bar A_{q_i} =
  \begin{pmatrix}
    0 & D_x l_{q_i} \\
    0 & A_{q_i}
  \end{pmatrix}\bigg |_{(x_n,u_n)}\text{\,.}
\end{displaymath}

\newer{In Mayer form, the first component of the perturbed appended state, $\bar x_{w,1}(t_f,\epsilon)$, is the perturbed integral cost in \eqref{hyb_J}.
Hence, the perturbed cost model \eqref{hyb_Jw} can be written as a sum, 
\[
J_w(\bar x_w,\epsilon) \triangleq \bar x_{w,1}(t_f,\epsilon) + m(x_w(t_f,\epsilon)) \text{\,,}
\]
which includes the perturbed terminal cost.
The direction of variation in the cost is $D_\epsilon J_w(\bar x_w,0) \triangleq \nu(t_f) = \bar \Psi_1(t_f) + D_\epsilon m(x_w(t_f,0))$.
Evaluating the derivative yields
\setlength{\arraycolsep}{0.0em}
\begin{eqnarray}
\label{hyb_nu}
\nu(t_f){ }={ } &&\bar \Psi_1(t_f) + D_x m(x(t_f)) \Psi(t_f) \nonumber\\
{ }={ } && [1, \nabla m(x(t_f))] \cdot \bar \Psi(t_f) \text{\,.}
\end{eqnarray}
}\noindent
\newer{\indent Note that $\nu(t_f)$ provides the same information as the mode insertion gradient in \eqref{dJdlambda} but applies to hybrid impulsive systems with resets.  That is, $\nu(t_f)$ provides the sensitivity of a cost, $J$, to applying an action at $t=\tau$ as $\lambda \rightarrow 0^+$.
Given a control perturbation at arbitrary time $\tau \in (t_0,t_f)$, one can calculate $\nu(t_f)$ from \eqref{hyb_nu} by propagating the appended state variation forward from $t=\tau$ to $t=t_f$ using Prop.~\ref{prop:variational_eq}.
However, in searching for an optimal time to act, SAC needs to compare the cost variation produced by taking action, i.e., applying a control perturbation, at different times, $\tau \in (t_0, t_f)$ (see Sec.~\ref{search}).
The process is computationally intensive if $\nu(t_f)$ is naively computed from the state variation. That is,
considering two possible times, $\tau < \tau'$, when control perturbation may be applied, $\nu(t_f)$ would require separate simulations of $\bar \psi$ from $[\tau,t_f]$ and $[\tau',t_f]$.\footnote{\new{One may also apply linear transformations to the variational system simulated from the perturbation at $\tau$ based on superposition of the initial condition at $\tau'$.  Variational reset maps would require similar transformation.}}}

\newer{
Since the mode insertion gradient \eqref{dJdlambda} does not require re-simulation to consider different application times $\tau \in (t_0,t_f)$ in optimizing \eqref{fcost}, we seek to express $\nu(t_f)$ in a form that more closely resembles \eqref{dJdlambda}.  To these ends, we now re-write $\nu(t_f)$ in terms of}
an \emph{adjoint} system, $\bar \rho$,\footnote{\new{The adjoint belongs to the cotangent bundle, $\bar \rho \in  T^*\mathcal{M}_{q_i}$, such that $\bar \rho(t) : T_x \mathcal{M}_{q_i} \mapsto \reals{}, \;\forall t \in \mathcal{I}_{q_i}, \forall  q_i \in q$.}} to the variational system $\bar \Psi$.\footnote{\new{See \cite{liberzonOptControl2012} for a similar derivation of an adjoint in the context of continuous variations.}}  The systems are adjoint \cite{liberzonOptControl2012} if 
\begin{equation}
\label{hyb_adjoint}
\frac{d}{dt}(\bar \rho \cdot \bar \Psi) = 0 = \dot{\bar \rho} \cdot \bar \Psi + \bar \rho \cdot \dot{\bar \Psi} \text{\,.} 
\end{equation}
That is, \newer{we can derive $\bar \rho$ by ensuring $\bar \rho \cdot \bar \Psi$ is constant.
Note also that by choosing the terminal condition, 
\begin{equation}
\label{hyb_adjoint_tf}
\bar \rho(t_f) = [\,1,\, \nabla m(x(t_f))\,] ,
\end{equation}
\eqref{hyb_nu} allows us to express $\nu(t_f)$ in terms of the adjoint at the terminal time as} $\nu(t_f) = \bar \rho(t_f) \cdot \bar \Psi(t_f)$.
\newer{If we enforce \eqref{hyb_adjoint} in deriving the adjoint}, the inner product will be constant and equal to $\nu(t_f)$ \edit{at times subsequent to the control perturbation, $\bar \rho(t) \cdot \bar \Psi(t) = \nu(t_f) \;\forall t \in [\tau,t_f]$, \newer{and $\bar \rho(t)$ can be interpreted as the sensitivity of \eqref{hyb_J} to a state variation at time $t$.}}

Assuming the system is at the (arbitrary) location $q \in \mathcal{Q}$ at the perturbation time $t = \tau$, the inner product in \eqref{hyb_adjoint} yields
\setlength{\arraycolsep}{0.0em}
\begin{eqnarray}
\label{hyb_mode_insert_grad}
\bar \rho(\tau) \cdot \bar \Psi(\tau) { }={ } && \bar \rho(\tau) \cdot \bigg ( \bar f_{q}(x_n(\tau),w) { }-{ } \bar f_{q}(x_n(\tau),u_n(\tau)) \bigg ) a\nonumber\\
{ }={ }&& \nu(t_f) \text{\,,}%
\end{eqnarray}
\newer{which no longer depends on forward simulation of $\bar \Psi(\tau)$.}
\edit{The initial time, $\tau$, of the control perturbation is arbitrary.
\newer{Like in \eqref{dJdlambda}, once the adjoint, $\bar \rho$, is computed over $[t_0,t_f]$, \eqref{hyb_mode_insert_grad} can be evaluated at any number of different times, $\tau \in (t_0,t_f)$, to provide the cost sensitivity, $\nu(t_f)$, to the control perturbation in each case.}}

\newer{The following proposition derives an adjoint formula, $\bar \rho$, that maintains its interpretation as the cost sensitivity to state variations, as in $\bar \rho(t) \cdot \bar \Psi(t) = \nu(t_f)\;\forall t \in [\tau,t_f]$.}
\begin{proposition}
\label{prop:adjoint_eq}
Assuming $x_n$ flows between the locations $q_i,q_{i+1} \in \mathcal{Q}$ with a control perturbation as in Prop.~\ref{prop:variational_eq},
\setlength{\arraycolsep}{5pt}
\begin{equation}
\label{hyb_rho_dot_complex}
   \bar \rho \triangleq \left\{
     \begin{array}{ll}
       [\,1,\, \nabla m(x(t_f))\,] & : t = t_f\\
       \dot{\bar \rho} = - \bar A_{q_{i+1}}^T \bar \rho &  : t \in [t_i^+, t_f)\\
       \bar \rho(t_i^-) = \bar \Pi_{q_i,q_{i+1}}^T \bar \rho(t_i^+) &  : t = t_i^-\\
       \dot{\bar \rho} = - \bar A_{q_{i}}^T \bar \rho &  : t \in [\tau, t_i^-)
     \end{array}
   \right.\text{\,,}
\end{equation}
satisfies the adjoint relation \eqref{hyb_adjoint} and \eqref{hyb_mode_insert_grad}.
\end{proposition} 
\begin{IEEEproof}
The adjoint is simulated backwards from a terminal condition \eqref{hyb_rho_dot_complex} because this choice of terminal conditions yields $\nu(t_f)$ in \eqref{hyb_nu}.  The continuous flow equations in rows 2 and 4 of \eqref{hyb_rho_dot_complex} are the direct result of enforcing \eqref{hyb_adjoint} with rows 2 and 4 of \eqref{hyb_psi_dot_complex}.  Similarly, the reset equation results from application of the adjoint relation across the transition time,
\setlength{\arraycolsep}{0.0em}
\begin{eqnarray}
\frac{d(\bar \rho(t_i) \cdot \bar \Psi(t_i))}{dt} { }={ } &&0 { } = { } \frac{\bar \rho(t_i^+) \cdot \bar \Psi(t_i^+)-\bar \rho(t_i^-) \cdot \bar \Psi(t_i^-)}{t_i^+ - t_i^-}\nonumber\\
{ } &&0{ }={ } \bar \rho(t_i^+) \cdot \bar \Pi_{q_i,q_{i+1}} \bar \Psi(t_i^-) - \bar \rho(t_i^-) \cdot \bar \Psi(t_i^-) \nonumber\\
\bar \rho(t_i^-) { }={ } &&\bar \Pi_{q_i,q_{i+1}}^T \bar \rho(t_i^+)\text{\,.}\nonumber
\end{eqnarray}
\end{IEEEproof}

\newer{As for the variational equation, one may propagate $\bar \rho$ between arbitrary numbers of consecutive modes by repeating the reset and continuous flow steps in \eqref{hyb_rho_dot_complex}.}

\newer{When the incremental costs, $l$, do not depend on the control, e.g., in \eqref{J}, and $w$ corresponds to the value of an optimal SAC action, \eqref{hyb_mode_insert_grad} is equivalent to the mode insertion gradient \eqref{dJdlambda}.\footnote{\newer{Appendix~\ref{sec:mode_insert_grad} details the connection between \eqref{dJdlambda} and \eqref{hyb_mode_insert_grad}.  The section also describes how \eqref{hyb_mode_insert_grad} can consider dynamic modes that differ in more than control, to enable mode scheduling algorithms for more general classes of hybrid systems with resets.}}  Hence, the SAC process applies as-is to hybrid and impulsive systems.  The user need only account for hybrid transitions in simulations, e.g., of state trajectory and the adjoint \eqref{hyb_rho_dot_complex}.}

\section{Hybrid Control Examples}
\label{sec:hyb_examps}

This section presents three illustrative examples using the hybrid methods just described. Section~\ref{sec:1d_bounce_ex} demonstrates calculation of the variational, adjoint, and \emph{hybrid mode insertion gradient} \eqref{hyb_mode_insert_grad} equations for a 1D example.  Section~\ref{sec:ball_ex} uses the hybrid version of SAC (based on the adjoint in \eqref{hyb_rho_dot_complex}) to control a bouncing ball through impacts and toward a goal state.  Lastly, Sec.~\ref{sec:slip_ex} applies SAC to control a the hybrid spring-loaded inverted pendulum model up a flight of stairs.

\subsection{Variations, Adjoint, and Control Sensitivity for a 1D Bouncing Mass}
\label{sec:1d_bounce_ex}

\begin{figure}[t!]
\centering
  \includegraphics[width=3.0in]{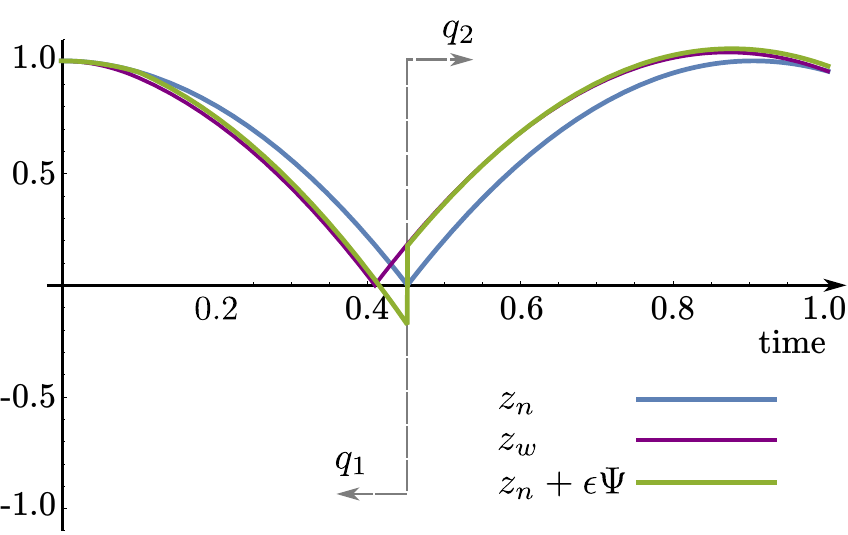}
  \caption{\new{The height $z(t)$ of a mass dropped from $1\,$m.  The mass follows the nominal trajectory, $z_n$, and bounces at impact due to an elastic collision.  The reset map reflects its velocity in transitioning from $q_1$ to $q_2$.  The purple curve is the varied trajectory simulated from the hybrid impulsive dynamics with a variation at $\tau = 0.1\,$s of duration $\lambda = 0.1\,$s ($a = 1$, $\epsilon = 0.1\,$s), in the nominal control.  The variation accelerates the mass in the $z$ direction at $w = -5 \,\frac{\text{m}}{\text{s}^2}$.  The green curve is the approximated trajectory based on the first-order model.}}
\label{fig:ex_simple_traj}
\end{figure}

This section computes variational and adjoint equations for a simple point mass system with impacts. The point mass is released from a height of $z_0 = 1\,$m with no initial velocity.  It falls under gravity until it impacts with a flat surface (guard) at $z = 0\,$m.  The dynamics before and after impact (\hl{locations} $q_1$ and $q_{2}$, respectively) are the same, corresponding to a point mass in gravity.  
However, a reset map reflects $\dot z$ when the guard becomes $0$ at impact.  The simulation parameters follow.

\begin{center}
\vspace{-.15in}
\begin{tabular}{lp{1.7in}}
\textbf{System Parameters:} & \\[2pt]
$\vec{x} = (\,z,\, \dot z \,)$ & $\vec{f}_{q_1}(\vec x,u) = (\, \dot z,\, -g + u \,)$ \\
$g = 9.81\,\frac{\text{m}}{\text{s}^2}$ & $\vec{f}_{q_2}(\vec x,\vec u) = \vec{f}_{q_1}(\vec x,\vec u)$ \\
$J = \int_{t_0}^{t_f} \vec{x}^T Q \,\vec x \,dt$ & $\mat{Q} = \mat{Diag}[\, 200,\, 0.01 \,]$ \\
$\mat \Omega_{q_1,q_2}(\vec x) = \mat{Diag}[\, 1,\, -1 ]\, \vec x$ & $\mat \Phi_{q_1,q_2}(x) = z$%
\end{tabular}
\end{center}

\begin{center}
\begin{tabular}{lp{1.9in}}
\textbf{Control Perturbation:} &  \\[2pt]
$\vec{u}_n = 0 \,\frac{\text{m}}{\text{s}^2}$ & $\tau = 0.1\,$s  \\
$a = 1$ & $\epsilon = 0.1\,$s \\
$w = -5 \,\frac{\text{m}}{\text{s}^2}$ & $\lambda = 0.1\,$s \\
\end{tabular}
\end{center}

\begin{figure}[t!]
\centering
  \includegraphics[width=3.0in]{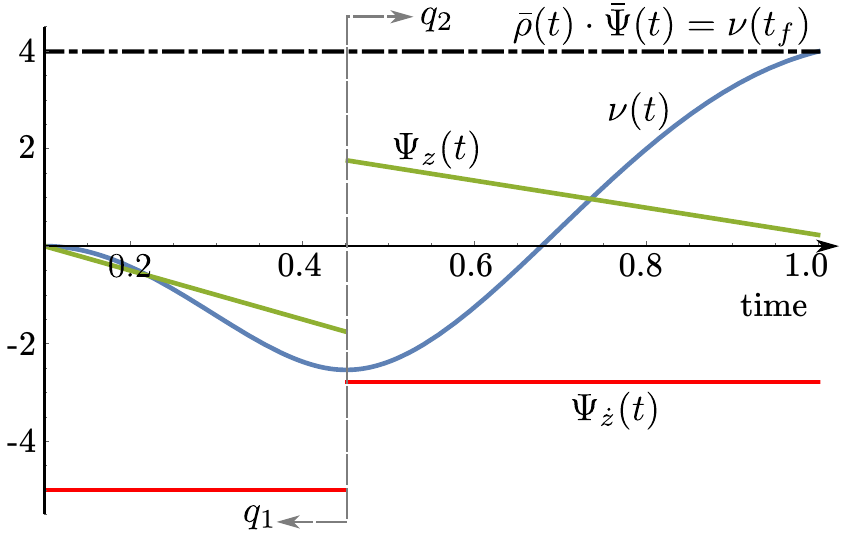}
  \caption{\new{The direction of state (and cost) variations, $\bar \Psi = (\,\nu,\, \Psi_z,\, \Psi_{\dot z}\,)$, resulting from a variation at $\tau = 0.1\,$s of duration $\lambda = 0.1\,$s ($a = 1$, $\epsilon = 0.1\,$s) in the nominal control.  The control variation accelerates the falling mass in the $z$ direction at $w = -5 \,\frac{\text{m}}{\text{s}^2}$. At all times subsequent the control variation, $\forall t \in [\tau, t_f]$, $\bar \rho(t) \cdot \bar \Psi(t)$ is equal to the direction of variation in the cost propagated to the final time, $\nu(t_f)$.  The state variations in $z$ and $\dot z$ are discontinuous at the transition from $q_1$ to $q_2$, while the cost variation is continuous.}}
\label{fig:ex_simple_variat_eq}
\end{figure}

Figure~\ref{fig:ex_simple_traj} shows the system's nominal trajectory (blue curve) and the varied trajectory resulting from a simulated 
\newer{control variation.}  The varied trajectory is computed from both the first-order variational model (green curve) and the true, nonlinear hybrid impulsive dynamics (purple curve).  
The variation directions resulting from \eqref{hyb_psi_dot_complex} are in Fig.~\ref{fig:ex_simple_variat_eq}.
As Fig.~\ref{fig:ex_simple_variat_eq} shows, the state variations are discontinuous at impact, while the direction of the cost variation, $\nu(t)$, is continuous over time.
The dashed black line in Fig.~\ref{fig:ex_simple_variat_eq} confirms the inner product, $\bar \rho \cdot \bar \Psi$, is constant and equal to the direction of the cost variation, $\nu(t_f)$, for all time subsequent the control perturbation, $\forall t \in [\tau,t_f]$.
Figure~\ref{fig:ex_simple_mode_insert} shows how this inner product (the value of \eqref{hyb_mode_insert_grad}) would change if the control perturbation were applied at different times, $\tau \in (t_0,t_f)$.  

\begin{center}
\vspace{-.2in}
\begin{tabular}{lp{1.35in}}
& \\
\textbf{\hl{Results:}} &  \\[2pt]
$\bar \rho(\tau) \cdot \bar \Psi(\tau) \big |_{\tau = 0.1}= 4$ & \\[5pt]
$\nu(t_f) = [\,1,\, 0,\, 0\,]^T \cdot \bar \Psi(t_f) = 4$ & \\[4pt]
$\Delta t_i \approx \epsilon \,\frac{d\Delta t_i}{d\epsilon} \big |_{\epsilon \rightarrow 0} = -0.04\,$s & \\[4pt]
$
  \Pi_{q_1,q_2} =
  \begin{pmatrix}
    -1 & 0 \\
    \frac{-2g+2u}{\dot z} & -1
  \end{pmatrix}
$ & \\
\end{tabular}
\end{center}

As asserted earlier, the approximation of the change in cost \eqref{hyb_J} from \eqref{hyb_mode_insert_grad} agrees with the first-order approximation of the change in cost from simulation of $\bar \Psi(t_f)$.
The first-order variational model, $z_n + \epsilon \Psi$, in Fig.~\ref{fig:ex_simple_traj} closely approximates the true perturbed trajectory, $z_w$, simulated from the perturbed control and the nonlinear dynamics.  Additionally, \eqref{hyb_dtideps} estimates the impact time of the varied system as $t = 0.41\,$s, which is near the updated impact time of $z_w$ in Fig.~\ref{fig:ex_simple_traj}.  Figure~\ref{fig:ex_simple_mode_insert} shows that \eqref{hyb_mode_insert_grad} correctly indicates it will be helpful (reduce trajectory cost according to \eqref{hyb_J}) to apply the control perturbation (push the mass toward the ground) after impact, when the ball is moving away from the ground.  Similarly, the figure suggests it will be detrimental to apply the control perturbation before impact because it would result in a net gain (positive change) in trajectory cost according to the first-order model.  %

\begin{figure}[t!]
\centering
  \includegraphics[width=3.0in]{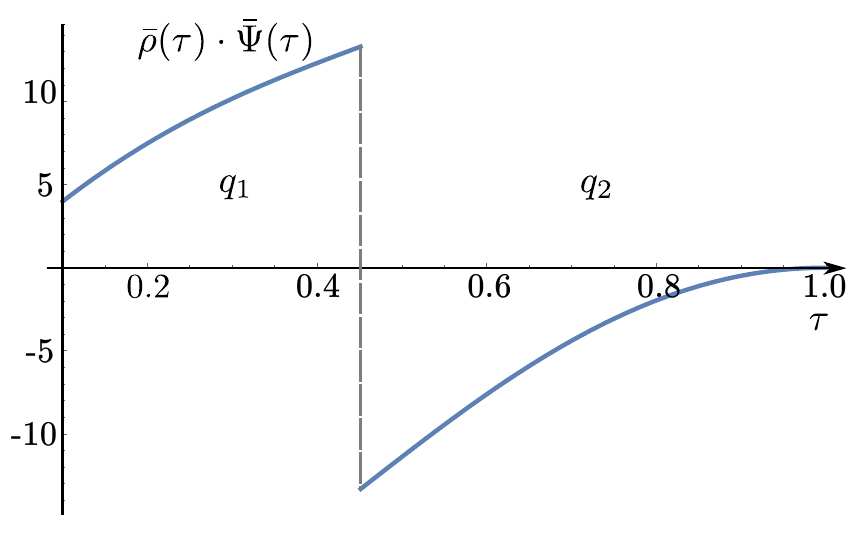}
  \caption{\new{The value of $\bar \rho(\tau) \cdot \bar \Psi(\tau)$ (according to \eqref{hyb_mode_insert_grad}) versus $\tau$.  The term indicates the sensitivity of the performance objective to the control perturbation, $w = -5 \,\frac{\text{m}}{\text{s}^2}$, if that perturbation were to occur at different points $\tau \in (t_0,t_f)$.  Before impact, \eqref{hyb_mode_insert_grad} indicates a short control perturbation will increase cost \eqref{hyb_J}.  %
After impact, a short control perturbation will lower cost.
}}
\label{fig:ex_simple_mode_insert}
\end{figure}

Finally, note that the reset map, $\Pi_{q_1,q_2}$, is only defined for velocities $\dot z$ that are non-zero.  As is typical for hybrid systems, these methods require that some component of the system's velocity vector lie in the direction of the switching surface so as to preclude grazing impacts.  The requirement ensures both \eqref{hyb_dtideps} and \eqref{hyb_pi} are well defined with $D_x\Phi_{q,q'}(x_n(t_i^-))f_{q_i}(x_n(t_i^-),u_n(t_i^-)) \neq 0 \; \forall (q,q') \in \mathcal{Q}$.

\begin{figure*}[t!]
\begin{subfigure}[b]{.48\textwidth}
\centering
\includegraphics[width=.8\textwidth]{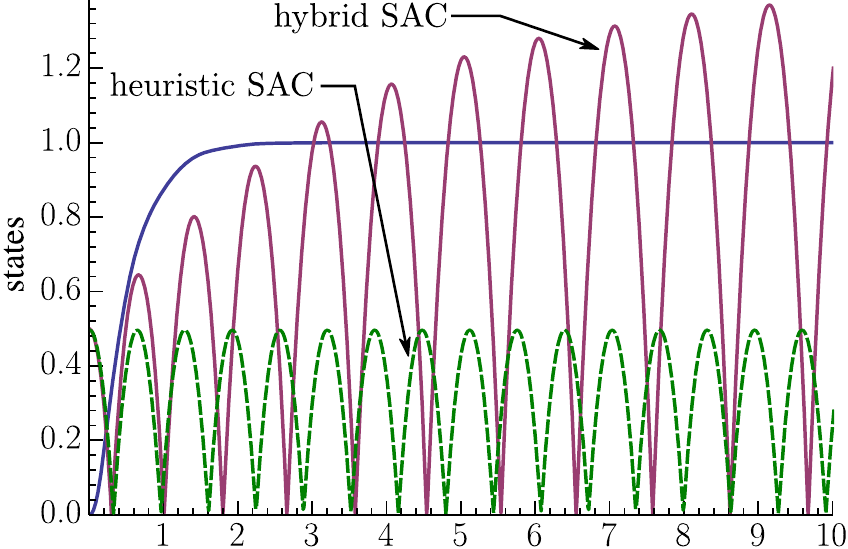}
\includegraphics[width=.8\textwidth]{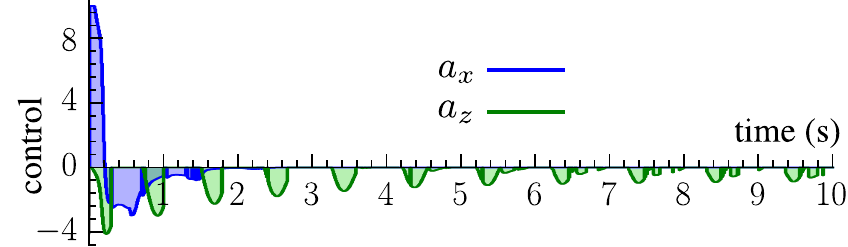}
\caption{ }
\label{fig:bounce_up}
\end{subfigure}
\hspace{.1in}
\begin{subfigure}[b]{.48\textwidth}
\centering
\includegraphics[width=.8\textwidth]{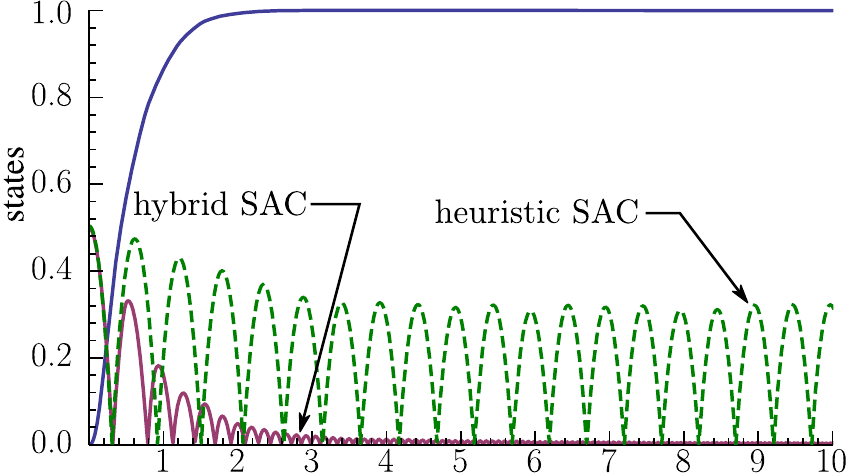}
\includegraphics[width=.8\textwidth]{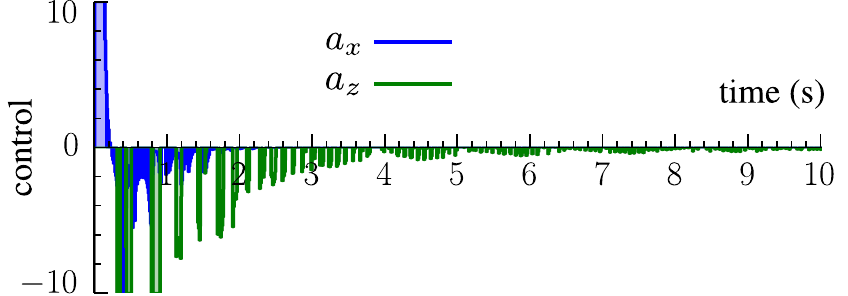}
\caption{ }
\label{fig:bounce_down}
\end{subfigure}
\caption{
\new{SAC accelerates a ball $1\,$m to the right and either up (Fig.~\ref{fig:bounce_up}) or down (Fig.~\ref{fig:bounce_down}).  In both cases $x_b$ (the blue state curve) reaches the desired point $1\,$m away.  In Fig.~\ref{fig:bounce_up}, control constraints prohibit the ball from accelerating against gravity, and so it cannot come to rest at the desired height. Instead, SAC accelerates the ball into the floor to rebound, increasing its height, $z_b$ (purple state curve), to maximize the time spent around the desired height of $1\,$m.  If the smooth version of SAC is applied as a heuristic (without the hybrid modifications), SAC drives the ball to the desired horizontal position but will not thrust in the $a_z$ direction.  Hence, the ball will continuously bounce at the initial height.  Similarly, in Fig.~\ref{fig:bounce_down}, the hybrid version of SAC successfully reduces energy from the (conservative) system by accelerating the ball into the floor when its momentum is away from the floor. Though it gets indistinguishably close, the ball cannot come to rest on the ground or it would result in infinite switching.  If the smooth version of SAC is applied as a heuristic, SAC will drive the ball to the desired horizontal position but cannot reduce the bouncing height below $z_b \approx 0.3\,$m.}
}\label{fig:bounce_up_down}
\end{figure*}

\subsection{Control of A Bouncing Ball}
\label{sec:ball_ex}

This section uses %
the SAC algorithm with the adjoint variable \eqref{hyb_rho_dot_complex}\footnote{\new{The first term of $\bar \rho$ is always $1$ and can be stripped to obtain an unappended hybrid adjoint, $\rho$, which applies to unappended dynamics as in \eqref{dJdlambda} when the incremental cost does not depend on the control (as in \eqref{J}).}} to develop closed-loop controls on-line that drive a hybrid impulsive bouncing ball model toward different desired states.  The system state vector consists of the 2D position and velocity of the ball, $x = (\, x_b,\, z_b,\, \dot x_b,\, \dot z_b \,)$.  The system inputs are constrained accelerations, $u = (\, a_x,\, a_z\,) : a_x \in [-10,10\,]\,\frac{\text{m}}{\text{s}^2}, \,a_z \in [-10,0\,]\,\frac{\text{m}}{\text{s}^2}$, and the dynamics are \newer{$f_q(x,u) = (\, \dot x_b,\, \dot z_b,\, a_x,\, a_z-g \,) : \forall q \in \mathcal{Q}$.}
As in the previous example, impacts are conservative and so reflect velocity orthogonal to the surface.

The SAC algorithm is initialized from half a meter off the ground, $x_{init} = (\, 0,\, 0.5\,\text{m},\, 0,\, 0 \,)$, and results are presented for two different tracking scenarios assuming a flat floor at $z_b = 0\,$m as the impact surface (guard).  In the first case, SAC uses the quadratic tracking cost \eqref{Jquad} with $Q = Diag[\,0,\, 10,\, 0,\, 0 \,]$, $P = Diag[\,10,\, 0,\, 0,\, 0 \,]$, and applies $R = Diag[\, 1,\, 1 \,]$ with $T = 0.5\,$s, $\gamma = -10$, and feedback sampling at $100\,$Hz.\footnote{\new{The hybrid examples specify SAC with parameters that cause it to skip the (optional) control search process in Section~\ref{search} as it is unnecessary in these cases and complicates analysis.}}  In this scenario, SAC is set to minimize error between the trajectory of the ball and a desired state a meter to the right of its starting position and one meter above the ground, $x_d = (\, 1\,\text{m},\, 1\,\text{m},\, 0,\, 0 \,)$.  The $10\,$s closed-loop tracking results included in Fig.~\ref{fig:bounce_up} require $0.21\,$s to simulate using the C++ SAC implementation from Section~\ref{examples}.

Accelerating the ball in the horizontal directions, SAC drives the ball toward the desired horizontal position $1\,$m away.  Due to the control constraints on $a_z$, however, SAC cannot achieve the desired height.  Instead, Fig.~\ref{fig:bounce_up} shows SAC accelerates the ball into the ground to increase its height after impact.  The behavior cannot be achieved without the hybrid modifications to the adjoint variable \eqref{hyb_rho_dot_complex} introduced here.  
Without the jump terms in the adjoint simulation (from reset map $\Pi_{q,q'}$), the mode insertion gradient \eqref{dJdlambda} does not switch signs at impact events as in Fig.~\ref{fig:ex_simple_mode_insert} and so does not accurately model the sensitivity to control actions.  

A similar demonstration in Fig.~\ref{fig:bounce_down} shows SAC tracking the desired state, $x_d = (\, 1\,\text{m},\, 0,\, 0,\, 0 \,)$, which is also $1\,$m from the starting position but on the ground.  Results take $0.29\,$s to compute and are based on all the same parameters previously mentioned but with $Q = Diag[\,0,\, 0,\, 0,\, 10 \,]$, so that the cost includes errors on horizontal position (from the $P$ matrix specifying the terminal cost) and vertical velocity.  

Because the system is conservative, SAC must act in the $a_z$ direction to remove energy.  As SAC can only accelerate the ball into the ground, the algorithm waits until the ball's momentum carries it upward and away from the floor to apply control, $a_z$.  
Lastly, if one applies the smooth version of SAC from Section~\ref{control} to this control scenario, the algorithm will control the ball to the desired horizontal point. While it will reduce the height to approximately $0.3\,$m, it ceases to make further progress (see Fig.~\ref{fig:bounce_down}).  %
These findings highlight the fact that %
 \eqref{dJdlambda} provides a poor model for hybrid systems with many switching events.  
\begin{figure}[t!]
  \centering
  \includegraphics[width=2.3in]{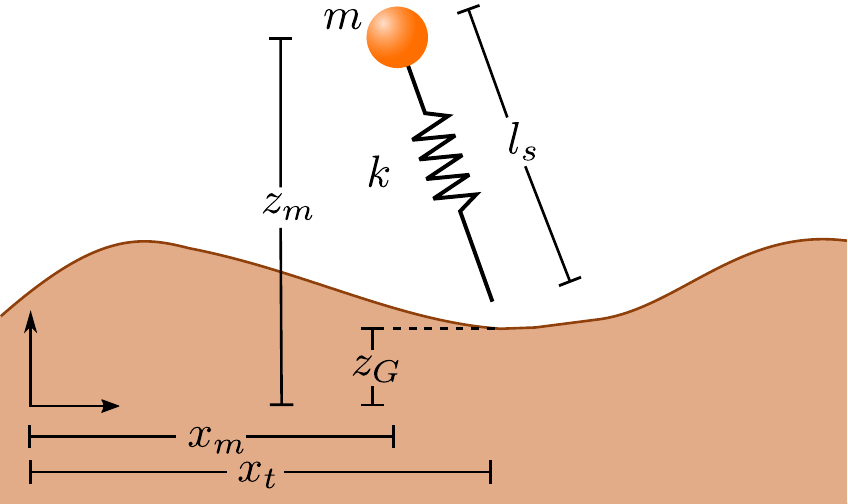}
  \caption{\new{Planar configuration variables for the SLIP.}}
\label{fig:slip_config}
\end{figure}

\subsection{Control of a Spring-Loaded Inverted Pendulum}
\label{sec:slip_ex}

This final example considers control for the \newer{SLIP model from the introduction.}
This section uses a 12 dimensional (9 states and 3 controls) model that is similar to the one in \cite{SLIPmodelControl2009}.  Figure~\ref{fig:slip_config} depicts the SLIP's planar configuration variables.

\begin{figure*}[t]
\begin{subfigure}[t]{.35\textwidth}
\includegraphics[width=3.2in]{figures/hopping_stairs_pt5sec_time_lapse_dwnSampled}
\caption{ }
\label{fig:slip_stairs}
\end{subfigure}
\hspace{.9in}
\begin{subfigure}[b]{.45\textwidth}
\centering
\includegraphics[width=3.5in]{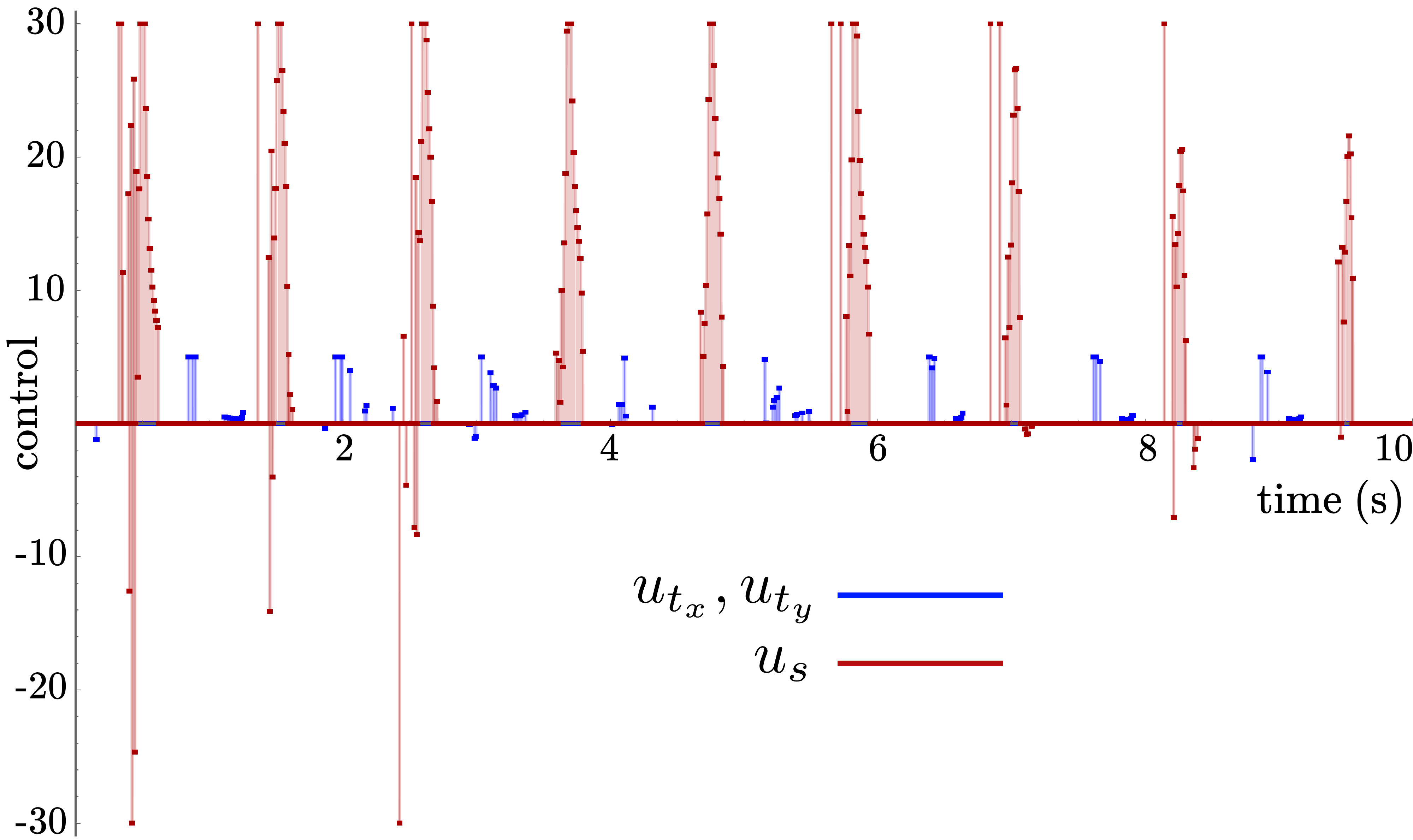}
\caption{ }
\label{fig:slip_stairs_control}
\end{subfigure}
\caption{\new{A time lapse showing the SLIP at $0.5\,$s increments (Fig.~\ref{fig:slip_stairs}) under SAC controls (Fig.~\ref{fig:slip_stairs_control}).}}\label{fig:slip_hopping}
\end{figure*}

The SLIP's dynamics are divided into flight and stance modes.  In our case, the state vector is the same for each mode and includes the 3D position / velocity of the mass, the 2D position of the spring endpoint (``toe''), and a \edit{bookkeeping variable, \newer{$q \in \{\mathrm{f},\mathrm{s}\}$}, tracking} the current hybrid location (indicating if the SLIP is in flight or stance), $\vec x = (\, x_{m}, \dot{x}_{m}, y_{m}, \dot{y}_{m}, z_{m}, \dot{z}_{m}, x_t, y_t, q \,)$.  The control vector is 3 dimensional, $\vec u = ( \, u_{t_x}, u_{t_y}, \newer{u_s} \,)$, composed of toe velocity controls, which can only be applied in flight, and the leg thrust during stance. The controls are further constrained so the toe velocities are $\in [-5,5] \, \frac{\text{m}}{\text{s}}$ and $\abs{u_s} \leq 30\,$N.

Ignoring the location variable, $q$, the stance dynamics,
\begin{equation}
\label{fs}
\newer{\vec f_{\mathrm{s}}}(\vec x,\vec u) =  \begin{pmatrix}
			  \dot{x}_{m} \\
			  \frac{(k (l_0-l_s)+u_s)(x_{m}-x_t)}{m l_s} \\
                          \dot{y}_{m} \\
			  \frac{(k (l_0-l_s)+u_s)(y_{m}-y_t)}{m l_s} \\
			  \dot{z}_{m}  \\
			  \frac{(k (l_0-l_s)+u_s)(z_{m}-z_G)}{m l_s}-g\\
			  0\\
                          0
			 \end{pmatrix}\text{\,,}
\end{equation}
define the first hybrid mode, and flight dynamics, 
\begin{equation} 
\label{ff}
\newer{\vec f_{\mathrm{f}}}(\vec x,\vec u) = \big (\, \dot{x}_{m},\, 0,\, \dot{y}_{m},\, 0,\, \dot{z}_{m},\, -g,\, \dot{x}_{m}+u_{t_x},\, \dot{y}_{m}+u_{t_y} \,\big ) ,
\end{equation}
define the second.  These dynamics depend on gravity, $g$, mass, $m = 1\,$kg, spring constant, $k = 100 \frac{\text{N}}{\text{m}}$, the ground height at the toe location, \newer{$z_G$}, and the leg length during stance,
\begin{equation}
\label{l}
l_s = \sqrt{(x_{m}-x_t)^2+(y_{m}-y_t)^2+(z_{m}-z_G)^2} .
\end{equation}
When $l_s = l_0$, the guard equations,
\begin{equation}
\label{phi}
\newer{\Phi_{\mathrm{f},\mathrm{s}}(x) = \Phi_{\mathrm{s},\mathrm{f}}(x)} =  z_m - \frac{l_0 (z_{m}-z_G)}{l_s} - z_G \text{\,,}
\end{equation}
cross zero to indicate the transition from stance to flight mode (and vice versa).  Upon transitioning to flight, the leg length becomes fixed at the resting length, $l_0 = 1\,$m.  Reset maps \newer{$\Omega_{\mathrm{f},\mathrm{s}}$ and $\Omega_{\mathrm{s},\mathrm{f}}$} leave the state unchanged other than to update the location variable, $q$.

Figure~\ref{fig:slip_hopping} includes a sample trajectory based on a quadratic objective with $\mat{Q} = \mat{Diag}[\,0\,, 70\,, 0\,, 70\,, 50\,, 0\,, 0\,, 0\,]$, $\mat{R} = \mat{I}$, $\mat{P}_1 = \mat{0}$, $T = 0.6\,$s, and $\alpha_d = -10$.
The figure depicts SAC controlling the SLIP up a staircase, which is approximating using logistic functions, $\displaystyle{\newer{z_G} = \sum_{\newer{n=1}}^4 \frac{0.5}{1 + e^{-75 (x - 0.7 \newer{n})}}}$.  
These functions produce stairs with a slope of $\approx 0.71$ (a $0.5\,$m rise every $0.7\,$m).  
As the rise of each step is equal to half the SLIP body length, SAC must coordinate leg motion to avoid tripping on the stair ledges.  
With the desired trajectory, $x_d = (\,0\,, 0.7 \,\frac{\text{m}}{\text{s}}\,, 0\,, 0.7 \,\frac{\text{m}}{\text{s}}\,, z_G + 1.4 \,\text{m}\,, 0\,, 0\,, 0\,)$, SAC drives the SLIP along a diagonal path up the staircase at roughly constant velocity and relatively uniform average height above the ground.  The $10\,$s trajectory simulates in $\approx 1.6\,$s on a laptop with feedback at $100$ Hz.\footnote{\new{The process is artificially slowed by impact event detection code, which we are still developing.}}  The hybrid SAC controller successfully navigates the SLIP over a variety of other terrain types, including sloped sinusoidal floors, using these same parameters and with similar timing results.  More recent results confirm SAC also extends to \newer{two-legged}, compliant walking models from \cite{SpringMassWalkRun2006}.  Both these varied terrain SLIP locomotion and compliant walking examples are in the video attachment.

The SLIP is well-studied, and researchers have already derived methods for stable hopping that control the SLIP leg to desired touchdown angles. 
These methods typically assume the leg can swing arbitrarily fast to implement analytically computed touchdown angles, ignoring
possible collisions with terrain during swing.
This example shows that the hybrid version of SAC can drive the SLIP over varying terrain while controlling the motion of the leg to avoid tripping.
We note that SAC implementations like the one introduced here may 
prove useful in controlling robots that (mechanically) emulate the SLIP \cite{ATRIASslipTemplate2014,AMASCslipActuator2010,RaibertHoppers1986}.
Due to physical constraints, these robots are limited in how well they can approximate the SLIP model assumptions and so SLIP-based control may prove ineffective.
In contrast, SAC can be applied to the actual robot model
(or a more accurate model) and does not rely on the simplifying SLIP
assumptions to control locomotion.

\end{newsection}

\section{Conclusions and Future Work}
\label{conclusion}

\begin{newsection}
This paper contributes a model-based algorithm, Sequential Action Control (SAC), that sequences optimal actions into a closed-loop, trajectory-constrained control at roughly the rate of simulation.  
\hl{While the approach is new and further study is required to define properties like robustness and sensitivities, we have tested SAC on an array of problems spanning several categories of traditionally challenging system types.}
These benchmark trials confirm the algorithm can outperform standard methods for nonlinear optimal control and case specific controllers in terms of tracking performance and speed.

For the continued development of SAC, a number of directions have been identified as possible future work.  
For instance, although we show SAC can avoid local minima that affect nonlinear trajectory optimization, the method is local in the sense that it cannot guarantee globally optimal solutions through state space (no method can in finite time for the nonlinear / non-convex problems here).
As such, despite the wide range of systems SAC can control, there are others that will prove difficult.  To increase applicability, SAC can be combined with global, sample-based planners to provide a fast local planner that develops constrained solutions that satisfy dynamics.  Such methods would allow SAC to apply even in very complicated scenarios such as those required to develop trajectories for humanoids \cite{Bretl3DWalking2012,iLQGmpcTodorov2012}.

To better automate policy generation and reduce required user input, SAC needs tools for parameter tuning, especially ones that provide stability.  As mentioned in Appendix~\ref{app:guarantees}, SAC parameters can be selected to provide local stability around equilibrium based on a linear state feedback law for optimal actions \eqref{ulinfb}.  Sums-of-Squares (SOS) tools, e.g., the S-procedure \cite{ParriloSOSPhd2000,LQRtreesTedrake2010}, seems a good candidate to automate parameter tuning and the generation of regions of attraction.

In addition to the applications mentioned, we note that SAC applies broadly to auto-pilot and stabilization systems like those in rockets, jets, helicopters, autonomous vehicles, and walking robots \cite{MPCvehicleStabilization2007,Bretl3DWalking2012,Helicopter2006,digitalFlightControl2008,KumarQuadrotors2012}.  It also naturally lends itself to shared control systems where the exchange between human and computer control can occur rapidly, e.g., wearable robotics and exoskeletons~\cite{BretlAssistLocomotion2010,lokomat2004,kazerooni2007,SACexosekApplic,alex2011}.  It offers a reactive, on-line control process that can reduce complexity and pre-computation required for robotic perching and aviation experiments in \cite{birdsFastFlyTrees2012,birdsFastFlyTreesICRA2014,LQRtreesTedrake2010}.
Its speed may facilitate predictive feedback control for new flexible robots~\cite{SoftOctopusArm2012,SoftRobot2011} and systems that are currently restricted to open-loop.  It offers real-time system ID and parameter estimation for nonlinear systems \cite{DOE2008,AndyOptParamEstimJour,SACParamEstim2015}.  These potential applications merit study and further development of the SAC approach.
\end{newsection}

\appendices
\section{}
\label{app:A}

\newer{The following sections highlight useful properties of SAC controls resulting from the synthesis process in Sec.~\ref{control}.}

\subsection{\newer{SAC Control Guarantees}}
\label{app:guarantees}

\begin{newsection}
\newer{Recall that SAC derives a schedule, $u_2^{\,*}$, that minimizes a convex objective \eqref{J2}, and causes the continuous first variation \eqref{deltaJ2} to vanish locally.  These are necessary and sufficient for the results in Corollary~\ref{global}.}

\begin{corollary}
\label{global}
\newer{Solutions $u_2^{\,*}$, in \eqref{U2Opt}, exist, are unique, and globally optimize the control cost, $J_2$, in \eqref{J2}.}
\end{corollary}

Additionally, the following Corollary~\ref{linfbprop} shows that
near equilibrium points, solutions \eqref{U2Opt} simplify to linear state feedback laws. This linear form permits local stability analysis (and parameter selection) based on continuous systems techniques.

\begin{corollary}
\label{linfbprop}
Assume system \eqref{f} is \newer{time invariant} with an equilibrium, $\newer{(x,u)} = 0$, the state tracking cost \eqref{J} is quadratic,\footnote{Quadratic cost \eqref{Jquad} is assumed so that resulting equations emphasize the local similarity between SAC controls and LQR \cite{AndersonMoore}.}
\begin{equation}
\label{Jquad}
J_1 { }={ } \frac{1}{2} \int_{t_0}^{t_f} \norm{\vec x(t) - \vec x_d(t)}_{\mat Q}^2 \,dt + \frac{1}{2} \, \norm{\vec x(t_f) - \vec x_d(t_f)}_{\mat{P_1}}^2\text{\,,}
\end{equation}
with $\vec x_d = \vec x_d(t_f) = \vec{0}$, $\mat Q = \mat{Q}^T \geq \mat 0$, and $\mat{P_1} = \mat{P_1}^T \geq \mat 0$, and $\vec u_1 = \vec 0$.  There exists \newer{a neighborhood, $\mathcal{N}(\newer{\vec x = \vec 0})$,}
where optimal actions \eqref{U2Opt} \newer{are linear} feedback regulators,
\begin{equation}
\label{ulinfb}
\vec u_2^{\,*}(t) = \alpha_d \, \mat{R}^{-1} \mat h(\new{0})^T \, \mat P(t) \, \vec x(t) \;\;\;\; \newer{\forall t \in (t_0,t_f)}\text{\,.}
\end{equation}
\end{corollary}

\begin{IEEEproof}
At the final time, $\vec \rho(t_f) = \mat{P_1} \vec x(t_f)$.  Due to continuity Assumps.~\ref{assump:dynam}-\ref{assump:cost}, this linear relationship must exist for a nonzero neighborhood of the final time, $\mathcal{N}(\newer{t = t_f})$, such that
\begin{equation}
\label{linearxrho}
\vec \rho(t) = \mat P(t) \, \vec x(t) \;\;\;\; \forall t \in \mathcal{N}(\newer{t=t_f})\text{\,.}
\end{equation}
Applying this relationship, \eqref{U2Opt} can formulated as
\setlength{\arraycolsep}{0.0em}
\begin{eqnarray}
\vec u_2^{\,*} { }=&{ }& \, ( \mat h(\vec x)^T \mat P \, \vec x \, \vec x^T \mat P^T \mat h(\vec x) + \mat{R}^T )^{-1} \nonumber\\
{ }&{ }& \, [ \mat h(\vec x)^T \mat P \, \vec x \, \vec x^T \mat P^T \mat h(\vec x) \, \vec u_1 + \mat h(\vec x)^T \mat P \, \vec x \, \alpha_d ]\text{\,.}\nonumber%
\end{eqnarray}
\setlength{\arraycolsep}{5pt}
\hspace{-4pt}This expression contains terms quadratic in $\vec x$.  For $\vec x \in \mathcal{N}(\newer{\vec x=\vec 0})$, these quadratic terms go to zero faster than the linear terms, and controls converge to \eqref{ulinfb}.

\newer{Near the equilibrium, the dynamics can be approximated to first order as $\dot{\vec x} \approx \mat A \, \vec x + \mat B \, \vec u$, with LTI linearizations $A=D_xf(0,0)$ and $B=D_uf(0,0)$.
Note the state in \eqref{linearxrho} is the nominal state from control, $u=u_1$, and is assumed in $\mathcal{N}(\newer{\vec x=\vec 0})$. Thus, when $u_1 \in \mathcal{N}(u=0)$, the system is near the equilibrium and \eqref{linearxrho} can be differentiated using the approximated dynamics (with $u=u_1$) and \eqref{rhodot} to show}
\setlength{\arraycolsep}{0.0em}
\begin{eqnarray}
\label{lyap_precursor}
\dot{ \vec \rho} { }=&{ }& \, \dot{ \mat P} \, \vec x + \mat P \, \dot{\vec x}\\
\newer{-Q x - \mat A^T \mat P \, \vec x} { }=&{ }& \, \dot{\mat P} \, \vec x + \mat P \, (\mat A \, \vec x + B \, u_1)\nonumber\text{\,.}%
\end{eqnarray}
\newer{When $\vec u_1 = \vec 0$, \eqref{lyap_precursor} reduces to}
\begin{equation}
\label{lyap}
\mat 0 = \mat Q + \mat A^T \mat P + \mat P \, \mat A + \dot{\mat P}\text{\,.}
\end{equation}
Note the similarity to a Lyapunov equation.  \newer{Though we have only proved this relationship exists in neighborhoods $\mathcal{N}(\newer{t=t_f})$ and $\mathcal{N}(\newer{\vec x = \vec 0})$, because \eqref{lyap} is linear in $P$, \eqref{lyap} cannot exhibit finite escape time.
Through a global version of the Picard--Lindel$\ddot{\text{o}}$f theorem \cite{nonlinSys2002}, it is straightforward to verify \eqref{lyap} (and \eqref{ulinfb}) exists and is unique for arbitrary horizons and not only for $t \in \mathcal{N}(\newer{t=t_f})$.
Hence, one can compute the time varying linear feedback regulators \eqref{ulinfb}\footnote{\new{Note the $h(0)^T = B^T$ term in \eqref{ulinfb} shows up because the system is assumed to be in a neighborhood where the dynamics can be linearly modeled.}} 
for $x \in \mathcal{N}(\newer{\vec x = \vec 0})$ with $\mat P(t)$ from \eqref{lyap} and $\mat P(t_f) = \mat{P_1}$. }
\end{IEEEproof}

\new{
\newer{Assuming time invariant dynamics, a fixed horizon, $T$, and} SAC continuously applies actions at the (receding) initial time, $t = t_0$, \eqref{ulinfb} yields a \emph{constant} feedback law, $\vec u_2^{\,*}(t) = -K \,\vec x(t)$, where $K$ depends on the linearizations, weights, $Q$, $R$, and $P_1$, the time horizon, $T$, and the $\alpha_d$ term.
Thus LTI stability conditions may be applied to facilitate parameter selection.\footnote{\newer{As an example,} Sums-of-Squares (SOS) \cite{ParriloSOSPhd2000,LQRtreesTedrake2010} techniques can pre-compute regions of attraction for \eqref{ulinfb}.  \newer{These SOS methods can be applied} for parameter optimization, or to determine when SAC should switch to continuous application of \eqref{ulinfb}.}}  Similarly, one can also show Corollary~\ref{linfbprop} yields a feedback expression in error coordinates for which LTV stability analysis can be used to identify parameters that guarantee local stability to a desired trajectory, $x_d(t)$.\footnote{\newer{In the LTV case, one would need to pre-compute the feedback matrix, $K(t)$, from each sample time when tracking a desired trajectory, $(x_d,u_d)$, in simulation.  Assuming fixing application times, e.g., $\tau=t_0$, for each action, one could interpolate between the constant feedback matrices to develop a single LTV feedback law for analysis, e.g., using SOS techniques \cite{ParriloSOSPhd2000,LQRtreesTedrake2010}.}}

\new{As a final point, if \eqref{J} is quadratic and the nominal control, $\vec u_1$,} modeled as applying consecutively computed optimal actions \eqref{ulinfb} near equilibrium, \eqref{lyap} becomes a Riccati differential equation for the closed-loop system (see \cite{LinSys2009}) and actions \eqref{ulinfb} simplify to finite horizon LQR controls \cite{AndersonMoore}.  In this case one can prove the existence of a Lyapunov function (\eqref{lyap} with $\dot{\mat P} = \mat 0$)
and guarantee stability for SAC using methods from LQR theory \cite{LinSys2009} to drive $\dot{\mat P} \rightarrow \mat 0$.  As for receding horizon control, \new{Lyapunov functions can be constructed} using infinite horizons or a terminal cost and constraints that approximate the infinite horizon cost \cite{NMPC2004,LyapFunMPC2011,NMPCbook2011,HauserTermCostStableNMPC2005,MPCreview2011,ConstraintStableMPC2000}.
\end{newsection}

\subsection{Input Constraints}
\label{app:saturation}

This section provides several means to incorporate min-max saturation constraints on elements of the optimal action vector.  To simplify the discussion and analysis presented, \newer{we assume $\vec u_1 = \vec{0}$, as in the implementation examples.}%

\subsubsection{Control Saturation -- Quadratic Programming}
\label{linProgSat}

While more efficient alternatives will be presented subsequently, the most general way to develop controls that obey saturation constraints is by minimizing \eqref{J2const} subject to inequality constraints.  The following proposition provides the resulting quadratic programming problem in the case of $\vec u_1 = \vec{0}$.

\begin{proposition}
\label{linprog}
At any application time $\tau$, a control action exists that obeys saturation constraints from the constrained quadratic programming problem
\begin{equation}
\label{J2const}
\vec u_2^{\,*}(\tau) = \argmin_{\vec u_2(\tau)} \; \frac{1}{2} \norm{ \vec \Gamma(\tau) \, \vec u_2(\tau) - \alpha_d}^2 + \frac{1}{2} \norm{ \vec u_2(\tau) }_{\mat R}^2
\end{equation}
such that $u_{min, k} \leq u_{2,k}^*(\tau) \leq \, u_{max, k} \;\forall k \in \{1, \dots , m\}$.
The term $\vec \Gamma^T \triangleq \mat h(\vec x)^T \, \vec \rho\;$, and values $u_{min, k}$ and $u_{max, k}$ bound the $k^{th}$ component of $\vec{u}_{2}^{\,*}(\tau)$.
\end{proposition}

\begin{IEEEproof}
For control-affine systems \newer{with $u_1 = 0$,} the mode insertion gradient \eqref{dJdlambda} simplifies to \newer{the inner product,}
\newer{\begin{equation}
\label{inner}
\frac{dJ_1}{d \lambda^+}(\tau,u_2^{\,*}(\tau)) = \langle \, \vec \Gamma(\tau)^T , \vec u_2^{\,*}(\tau) \, \rangle \text{\,.}
\end{equation}}\noindent
\indent With the linear mode insertion gradient \eqref{inner}, minimizing \eqref{J2const} subject to $u_{min, k} \leq u_{2,k}^*(\tau) \leq \, u_{max, k} \;\forall k$ is equivalent to optimizing \eqref{l2} at time $\tau$ to find a saturated action, $\vec u_2^{\,*}(\tau)$.
\end{IEEEproof}

Prop.~\ref{linprog} considers a constrained optimal action, $\vec u_2^{\,*}(\tau)$, at a fixed time.  However, the quadratic programming approach can be used to search for the schedule of solutions $\vec u_2^{\,*}$ that obey saturation constraints (though it would increase computational cost).  These quadratic programming problems can be solved much more efficiently than the nonlinear dynamics constrained programming problems that result when searching for finite duration optimal control solutions.  As described next, even the limited overhead imposed by these problems can be avoided by taking advantage of linearity in \eqref{U2Opt}.

\subsubsection{Control Saturation -- Vector Scaling}
\label{vecScalingSat}

Optimal actions computed from \eqref{U2Opt} are affine with respect to $\alpha_d$ and linear when $\vec u_1 = \vec{0}$.  Thus, scaling $\alpha_d$ to attempt more dramatic changes in cost relative to control duration produces actions that are scaled equivalently.\footnote{Generally, scaling $\alpha_d$ will not equivalently scale the overall change in cost because the neighborhood, $V$, where the \eqref{DeltaJ} models the change in cost can change.  This would result in a different duration $\lambda$ for the scaled action.}  
\newer{The} linear relationship between $\vec u_2^{\,*}(\tau)$ and $\alpha_d$ implies that if any component $u_{2,k}^{\,*}(\tau) > u_{max,k}$ or $u_{2,k}^{\,*}(\tau) < u_{min,k}$, one can choose a new $\hat \alpha_d$ that positively scales the entire control vector until constraints are satisfied.  If the worst constraint violation is due to a component $u_{2,k}^*(\tau) > u_{max,k}$, choosing $\hat \alpha_d = \alpha_d \, u_{max,k} / u_{2,k}^*(\tau)$ will produce a positively scaled $\vec u_2^{\,*}(\tau)$ that obeys all constraints.  Linearity between $\vec u_2^{\,*}(\tau)$ and $\alpha_d$ implies that this factor can be directly applied to control actions from \eqref{U2Opt} rather than re-calculating from $\hat \alpha_d$.  To guarantee that scaling control vectors successfully returns solutions that obey constraints and reduce cost \eqref{J}, constraints must be of the form $u_{min,k} < 0 < u_{max,k} \; \forall k$.
\begin{proposition}
\label{descent}
For the choice $\alpha_d < 0$, a control action $\vec u_2^{\,*}(\tau)$ evaluated anywhere that $\vec \Gamma(\tau)^T \triangleq \mat h(x(\tau))^T \, \vec \rho(\tau) \neq \vec{0} \in \reals{m}$ will result in a negative mode insertion gradient \eqref{dJdlambda} and so can reduce \eqref{J}.
\end{proposition}

\begin{IEEEproof}
Combining \eqref{U2Opt} with \eqref{inner}, optimal actions that reduce cost result in a mode insertion gradient satisfying
\setlength{\arraycolsep}{0.0em}
\begin{eqnarray}
\frac{dJ_1}{d \lambda^+}(\cdot,\cdot) { }=&{ }& \, \langle \, \vec \Gamma(\tau)^T , (\vec \Gamma(\tau)^T \vec \Gamma(\tau) + \mat{R}^T)^{-1} \, \vec \Gamma(\tau)^T \alpha_d  \, \rangle\nonumber\\
{ } = &{ }& \, \alpha_d \, \norm{\vec \Gamma(\tau)^T}_{(\vec \Gamma(\tau)^T \vec \Gamma(\tau) + \mat{R}^T)^{-1}}^2 \, < 0 \text{\,.}\nonumber%
\end{eqnarray}
\setlength{\arraycolsep}{5pt}
\hspace{-4pt}The outer product, $\vec \Gamma(\tau)^T \vec \Gamma(\tau)$, produces a positive semi-definite symmetric matrix.  Adding $\mat R > \mat 0$ yields a positive definite matrix.  Because the inverse of a positive definite matrix is positive definite, the quadratic norm $\norm{\vec \Gamma(\tau)^T}_{(\vec \Gamma(\tau)^T \vec \Gamma(\tau) + \mat{R}^T)^{-1}}^2 > 0$ for $\vec \Gamma(\tau)^T \neq \vec{0} \in \reals{m}$.  Therefore, only choices $\alpha_d < 0$ in \eqref{J2} produce optimal control actions that make $\frac{dJ_1}{d \lambda^+} < 0$ and by \eqref{DeltaJ} can reduce cost \eqref{J}.
\end{IEEEproof}

\subsubsection{Control Saturation -- Element Scaling}
\label{elemScalingSat}

\new{For} multidimensional vectors, scaling can produce overly conservative (unnecessarily small magnitude) controls when only a single vector component violates a constraint.  To avoid the issue and reap the computational benefits of vector scaling, one can choose to scale the individual components of a multi-dimensional action, $\vec u_2^{\,*}(\tau)$, by separate factors to provide admissible solutions (saturated control actions that reduce \eqref{J}).  The following proposition presents conditions under which this type of saturation guarantees admissible controls.

\begin{proposition}
\label{easySat}
Assume $\mat R = c \, \mat I$ where $\mat I$ is the identity and $c \in \reals{+}$, $\alpha_d \in \reals{-}$, $\vec u_1 = \vec 0$, and separate saturation constraints $u_{min,k} \leq 0 \leq u_{max,k} \; \forall k \in \{1, \dots , m\}$ apply to elements of the control vector.  The components of any control derived from \eqref{U2Opt} and evaluated at any time, $\tau$, where $\vec \Gamma(\tau)^T \triangleq \mat h(\vec x(\tau))^T \, \vec \rho(\tau) \neq \vec{0} \in \reals{m}$ can be independently saturated.  If $\norm{\vec u_2^{\,*}(\tau)} \neq 0$ after saturation, the action is guaranteed to be capable of reducing cost \eqref{J}.
\end{proposition}

\begin{IEEEproof}
For the assumptions stated in Prop.~\ref{easySat},
\[\vec u_2^{\,*}(\tau) = (\vec \Gamma(\tau)^T \vec \Gamma(\tau) + \mat{R}^T)^{-1} \, \vec \Gamma(\tau)^T \alpha_d \text{\,.}\]
The outer product, $\vec \Gamma(\tau)^T \vec \Gamma(\tau)$, produces a rank 1 positive semi-definite, symmetric matrix with non-zero eigenvalue $= \vec \Gamma(\tau) \vec \Gamma(\tau)^T$ associated with eigenvector $\vec \Gamma(\tau)^T$.  Eigenvalue decomposition of the outer product yields $\vec \Gamma(\tau)^T \vec \Gamma(\tau) = \mat S \, \mat D \, \mat{S}^{-1}$, where the columns of $\mat S$ corresponds to the eigenvectors of $\vec \Gamma(\tau)^T \vec \Gamma(\tau)$ and $\mat D$ is a diagonal matrix of eigenvalues.  For $\mat R = \mat{R}^T = c \, \mat I$, actions satisfy
\newer{\setlength{\arraycolsep}{0.0em}
\begin{eqnarray*}
\vec u_2^{\,*}(\tau) { } = &{ }& \, (\mat S \, \mat D \, \mat S^{-1} + c \, \mat I)^{-1} \, \vec \Gamma(\tau)^T \alpha_d\\
{ } = &{ }& \, (\mat S \, \mat D \, \mat S^{-1} + c \, \mat S \, \mat I \, \mat S^{-1})^{-1} \, \vec \Gamma(\tau)^T \alpha_d\\
{ } = &{ }& \, \mat S \, (\mat D + c \, \mat I)^{-1} \, \mat S^{-1} \, \vec \Gamma(\tau)^T \alpha_d \text{\,.}%
\end{eqnarray*}
\setlength{\arraycolsep}{5pt}}\noindent
\indent The matrix $\mat D + c \, \mat I$ must be symmetric and positive-definite with eigenvalues all equal to $c$ except for the one associated with the nonzero eigenvalue of $\mat D$.  This eigenvalue, $\vec \Gamma(\tau) \vec \Gamma(\tau)^T + c$, applies to eigenvectors that are scalar multiples of $\vec \Gamma(\tau)^T$.  After inversion, $\mat S \, (\mat D + c \, \mat I)^{-1} \, \mat S^{-1}$ must then have an eigenvalue $\frac{1}{\vec \Gamma(\tau) \vec \Gamma(\tau)^T + c}$.  Since inversion of a diagonal matrix leaves its eigenvectors unchanged, the eigenvalue scales $\Gamma(\tau)^T$.  Therefore, the matrix $\mat S \, (\mat D + c \, \mat I)^{-1} \, \mat S^{-1}$ directly scales its eigenvector, $\vec \Gamma(\tau)^T$, and
\begin{equation}
\label{U2Opteigen}
\vec u_2^{\,*}(\tau) = \frac{\alpha_d}{\vec \Gamma(\tau) \vec \Gamma(\tau)^T + c} \, \vec \Gamma(\tau)^T \text{\,.}
\end{equation}

For any $\alpha_d \in \reals{-}$, $\vec u_2^{\,*}(\tau)$ will be a negative scalar multiple of $\vec \Gamma(\tau)^T$. Because two vectors $\in \reals{m}$ can at most span a $2D$ plane $E \subset \reals{m}$, the Law of Cosines (the angle, $\phi$, between vectors $\vec u$ and $\vec v$ can be computed from $\cos(\phi) = \frac{\langle \vec u , \vec v \rangle}{\norm{\vec u} \norm{\vec v}}$) can be applied to compute the angle between any $\vec u_2^{\,*}(\tau)$ and $\vec \Gamma(\tau)^T$. The Law of Cosines verifies that control \eqref{U2Opteigen} is $180^{\circ}$ relative to $\vec \Gamma(\tau)^T$.  Therefore, \eqref{U2Opteigen} corresponds to the control of least Euclidean norm that minimizes \eqref{inner} and so maximizes the expected change in cost.  The Law of Cosines and \eqref{inner} also imply the existence of a hyperplane, $h_p := \{ \vec \nu(\tau) \in \reals{m} \, \vert \, \langle \, \vec \Gamma(\tau)^T , \vec \nu(\tau) \, \rangle = 0\}$, of control actions, $\vec \nu(\tau)$, orthogonal to both \eqref{U2Opteigen} and $\vec \Gamma(\tau)^T$.  This hyperplane divides $R^{m}$ into subspaces composed of vectors capable of reducing cost \eqref{J} (they produce a negative mode insertion gradient based on inner product \eqref{inner}) and those that cannot.

To show that saturation returns a vector in the same subspace as \eqref{U2Opteigen}, one can define the control in terms of component magnitudes, $\vec a = (a_1, \dots , a_m)$, and signed orthonormal bases from $\reals{m}$, $\hat {\mat e} = (\hat{\vec e}_1, \dots, \hat{\vec e}_m)$, so that $\vec u_2^{\,*}(\tau) = \vec a \, \hat{\mat e}$.  The Law of Cosines confirms that $\vec u_2^{\,*}(\tau)$ can be written only in terms of components $a_k$ and signed basis vectors $\hat{\vec e}_k$ within acute angles of the control.  Briefly, the law indicates an $a_k$ cannot be associated with any basis, $\hat{\vec e}_k$, at $90^{\circ}$ of the control because it would require $\langle \, \hat{\vec e}_k, \vec u_2^{\,*}(\tau) \, \rangle = 0$, implying $a_k = 0$.  Similarly, an $a_k$ cannot be associated with an $\hat{\vec e}_k > 90^{\circ}$ relative to the control because this is equivalent to $\langle \, \hat{\vec e}_k, \vec u_2^{\,*}(\tau) \, \rangle < 0$, and leads to an $a_k < 0$ that contradicts definition. 

Because \eqref{U2Opteigen} is represented by positively scaled bases within $90^{\circ}$ of $\vec u_2^{\,*}(\tau)$, all these vectors must lie on the same side of $h_p$ as \eqref{U2Opteigen}.  This is also true of any vector produced by a non-negative linear combination of the components of $\vec u_2^{\,*}(\tau)$.  Since there always exists factors $\in [0, \infty)$, that can scale the elements of an action vector until they obey constraints $u_{min,k} \leq 0 \leq u_{max,k} \; \forall k \in \{1, \dots , m\}$, saturated versions of \eqref{U2Opteigen} will still be capable of reducing cost for $\norm{\vec u_2^{\*}(\tau)} \neq \vec 0$.
\end{IEEEproof}

\section{}
\label{app:B}

\newer{The following appendix sections apply to the hybrid version of SAC in Part II.  Specifically, Appendix~\ref{sec:variational_eq} derives the formula for state variations in Prop.~\ref{prop:variational_eq}. Appendix~\ref{sec:mode_insert_grad} describes how \eqref{hyb_mode_insert_grad} generalizes the mode insertion gradient and applies to mode scheduling problems for hybrid impulsive systems.}

\subsection{\newer{Computing the Varied State}}
\label{sec:variational_eq}

\newer{According to Prop.~\ref{prop:variational_eq}, assume a control perturbation occurs at location $q_i \in \mathcal{Q}$ at $t=\tau$. While the system remains in the same location (no hybrid transitions), \cite{liberzonOptControl2012} shows the direction of state variations along the continuous trajectory segment satisfies,
\setlength{\arraycolsep}{0.0em}
\begin{eqnarray}
\dot \Psi { }={ } &&A_{q_i} \Psi : t \in \mathcal{I}_{q_i}\text{\,,} \label{hyb_psi_dot}\\
\Psi(\tau) { }={ } &&\bigg ( f_{q_i}(x_n(\tau),w) - f_{q_i}(x_n(\tau),u_n(\tau)) \bigg ) a\text{\,,}\label{hyb_psi_init}%
\end{eqnarray}
with $A_{q_i}(t) \triangleq D_x f_{q_i}(x_n(t),u_n(t)) : t \in \mathcal{I}_{q_i}$.}

\newer{To propagate the varied state \eqref{hyb_xw} to a new location $q_{i+1}$, we apply the reset map, $\Omega_{q_i,q_{i+1}}$, as in}
\newer{
\setlength{\arraycolsep}{0.0em}
\begin{eqnarray}
\label{hyb_xw_int_complex}
x_w(t,\epsilon) { }={ } && \Omega_{q_i,q_{i+1}} \bigg ( x_w(t_i,\epsilon) + \int_{t_i}^{t_i+\Delta t_i^-} \hspace{-18pt}f_{q_i}(x_w(s,\epsilon),u_n(s)) ds \bigg )\nonumber\\
{ }&& + \int_{t_i+\Delta t_i^+}^{t} f_{q_{i+1}}(x_w(s,\epsilon),u_n(s)) ds \;\;\;: t \in \mathcal{I}_{q+1}\text{\,.}\nonumber\\
{ }&&%
\end{eqnarray}
}\noindent
\newer{Note the nominal state, $x_n$, transitions from $q_i$ to $q_{i+1}$ at $t_i$,} $\Delta t_i \triangleq \Delta t_i(\epsilon)$ is the change in transition time due to the state variation (see Fig.~\ref{fig:variations}), and a ``$-$'' or ``$+$'' superscript (as in $t_i+\Delta t_i^+$) indicates the time just before or after \newer{the transition.}  

\newer{To obtain the first-order variational equation for $\Psi$ at $q_{i+1}$ due to a control perturbation at $q_i$, we differentiate \eqref{hyb_xw_int_complex} as $\epsilon \rightarrow 0$ with $f_{q_i}(x_n(t_i^-),u_n(t_i^-)) \triangleq f_{q_i}^-$ and $f_{q_{i+1}}(x_n(t_i^+),u_n(t_i^+)) \triangleq f_{q_i}^+$, as in}
\newer{
\setlength{\arraycolsep}{0.0em}
\begin{eqnarray}
\label{hyb_psi_int_complex}
\Psi(t) { }={ } &&D_x \Omega_{q_i,q_{i+1}}(x_n(t_i^-)) \bigg [ \Psi(t_i^-) + \frac{d\Delta t_i}{d\epsilon} \bigg |_{\epsilon \rightarrow 0} f_{q_i}^- \bigg ] \\
{ }&&- \frac{d\Delta t_i}{d\epsilon} \bigg |_{\epsilon \rightarrow 0} f_{q_{i+1}}^+ + \int_{t_i^+}^{t} A_{q_{i+1}}(s) \Psi(s) ds \;\;\;: t \in \mathcal{I}_{q_{i+1}}\text{.}\nonumber
\end{eqnarray}
}\noindent
\newer{We compute $\frac{d\Delta t_i}{d\epsilon} \big |_{\epsilon \rightarrow 0}$} by locally enforcing the guard equation,
\begin{equation}
\label{hyb_guard}
\Phi_{q_i,q_{i+1}}(x_w(t_i+\Delta t_i^-,\epsilon)) = 0 \text{\,,}
\end{equation}
using the first-order Taylor expansion of \eqref{hyb_guard} around \newer{$\epsilon \rightarrow 0$.
Applying} $\Phi_{q_i,q_{i+1}}(x_n(t_i^-)) = 0$ in the expansion yields,
\newer{
\begin{equation}
\label{hyb_dtideps}
\frac{d\Delta t_i}{d\epsilon} \bigg |_{\epsilon \rightarrow 0} = -\frac{D_x\Phi_{q_i,q_{i+1}}(x_n(t_i^-))\Psi(t_i^-)}{D_x\Phi_{q_i,q_{i+1}}(x_n(t_i^-))f_{q_i}^-}\text{\,.}
\end{equation}
}\noindent
Finally, one can define the new reset term\newer{, $\Pi_{q_i,q_{i+1}}$, according to \eqref{hyb_pi},}
\newer{and use \eqref{hyb_psi_dot}, \eqref{hyb_psi_init} and \eqref{hyb_dtideps} to express \eqref{hyb_psi_int_complex} as,}
\newer{
\setlength{\arraycolsep}{0.0em}
\begin{eqnarray}
\label{hyb_psi_int_complex_2}
\Psi(t) { }={ } &&\Pi_{q_i,q_{i+1}} \Psi(t_i^-) + \int_{t_i^+}^{t} A_{q_{i+1}}(s) \Psi(s) ds \;\;\;\; : t \in \mathcal{I}_{q_{i+1}}\text{\,.}\nonumber%
\end{eqnarray}
}\noindent
Just as $\Omega_{q_i,q_{i+1}}$ resets the state in \eqref{hyb_xw_int_complex}, $\Pi_{q_i,q_{i+1}}$ provides a (linear) reset map \newer{that transitions $\Psi$ between} locations in \eqref{hyb_psi_int_complex_2}. 

\newer{Rather than calculate from \eqref{hyb_psi_int_complex_2}, one can compute $\Psi$ from the series of differential equations and resets in Prop.~\ref{prop:variational_eq}}.

\subsection{\newer{The Hybrid} Mode Insertion Gradient}
\label{sec:mode_insert_grad}

\newer{Part I of this paper makes use of the mode insertion gradient to locally model the changes in cost \eqref{J} generated by SAC actions.
As in Sec.~\ref{control}, the mode insertion gradient in hybrid systems literature \cite{TimSuffDescentModeSched2016,MagnustInsertionGradientTransitionTime,tomlinModeSchedAlg2010,MagnusInsertionGradientDeriv,MagnusInsertionGradient} applies to incremental costs, $l_1$,} that do not depend on the \newer{control.}
With this assumption, the mode insertion gradient provides the sensitivity of $J_1$ to insertion of a \newer{new dynamic mode, i.e., switching from some nominal mode, $f_{q\in \mathcal{Q}}$, to another mode of the same dimension, $f_{q'\in \mathcal{Q}}$,} for a short duration around $\lambda \rightarrow 0^+$.
In the case of SAC, the alternate dynamic modes differ only in \newer{control and} so result in the form of the mode insertion gradient in \eqref{dJdlambda} for smooth systems.
\newer{The expression \eqref{hyb_mode_insert_grad} is a generalization of \eqref{dJdlambda} that applies to hybrid impulsive systems and to costs \eqref{hyb_J}, which may depend on the control.}

Using the methods presented, it is straightforward to modify the initial condition of the variational equation \eqref{hyb_psi_init} to accommodate an arbitrary dynamic mode insertion, $f_{q'}$, rather than a control perturbation.  Note the formulas for the variational flow \eqref{hyb_psi_dot_complex} and its corresponding adjoint equation would remain unchanged.  In this case, \eqref{hyb_mode_insert_grad} becomes the more general form of the mode insertion gradient from hybrid systems literature (as it considers more than just control perturbations), but applies to broader classes of hybrid impulsive systems with resets.
Hence, the derivations and hybrid adjoint and mode insertion gradient calculations \eqref{hyb_mode_insert_grad} introduced \newer{in Part II} can enable mode scheduling algorithms like those in \cite{TimSuffDescentModeSched2016,tomlinModeSchedAlg2010,MagnusInsertionGradientDeriv,MagnusInsertionGradient} for these larger classes of hybrid impulsive systems.

\section*{ACKNOWLEDGMENT}

This material is based upon work supported by the National Science Foundation under Grant CMMI 1200321. Any opinions, findings, and conclusions or recommendations expressed in this material are those of the author(s) and do not necessarily reflect the views of the National Science Foundation.

\IEEEtriggeratref{58}

\bibliographystyle{IEEEtran}
\bibliography{IEEEabrv,bib_abrv}

\end{document}